\documentclass{article}

\usepackage[dblblindworkshop, final]{neurips_2026}
\workshoptitle{AI4Mat}
\usepackage{graphicx}%
\usepackage{multirow}%
\usepackage{amsmath,amssymb,amsfonts}%
\usepackage{amsthm}%
\usepackage{mathrsfs}%
\usepackage[title]{appendix}%
\usepackage{xcolor}%
\usepackage{textcomp}%
\usepackage{manyfoot}%
\usepackage{booktabs}%
\usepackage{algorithm}%
\usepackage{algorithmicx}%
\usepackage{algpseudocode}%
\usepackage{listings}%
\usepackage{booktabs}
\usepackage{array}
\usepackage{graphicx}
\usepackage{siunitx} %DY
\usepackage[version=3]{mhchem} % Formula subscripts using \ce{}
\usepackage{color-edits}
\addauthor{dy}{purple}
\addauthor{pb}{blue}
\addauthor{pf}{red}
\addauthor{cl}{orange}
\newcommand{\bx}{\mathbf{x}}

\newcommand{\by}{\mathbf{y}}
\newcommand{\bh}{\mathbf{h}}
\usepackage{graphicx}
\usepackage{tikz}
\usepackage{amsmath,amssymb,bm}
\usepackage[mode=buildnew]{standalone}
\usetikzlibrary{
    arrows.meta,
    calc,
    positioning,
    shapes.geometric,
    shapes.misc
}

\usepackage[utf8]{inputenc} % allow utf-8 input
\usepackage[T1]{fontenc}    % use 8-bit T1 fonts
\usepackage{hyperref}       % hyperlinks
\usepackage{url}            % simple URL typesetting
\usepackage{booktabs}       % professional-quality tables
\usepackage{amsfonts}       % blackboard math symbols
\usepackage{nicefrac}       % compact symbols for 1/2, etc.
\usepackage{microtype}      % microtypography
\usepackage{xcolor}         % colors

\title{Scalable Bayesian Optimization of Composite Functions for Image-Based Inverse Problems in Materials Characterization}
\author{%
  Dasol Yoon$^{1,2}$\thanks{These authors contributed equally to this work.}%
  \hspace{0.15cm}\thanks{Corresponding authors: \texttt{dy327@cornell.edu},
    \texttt{david.a.muller@cornell.edu}, \texttt{pf98@cornell.edu}}
  \quad Poompol Buathong$^{3}$\footnotemark[1]
  \quad Chia-Hao Lee$^{1}$
  \quad Yujia Zhang$^{3}$ \\[0.25em]
  \textbf{David A.~Muller$^{1,4}$\footnotemark[2]
  \quad Peter I.~Frazier$^{5}$\footnotemark[2]} \\[0.7em]
  \normalfont\small
  \begin{tabular}{c}
    $^{1}$School of Applied and Engineering Physics, Cornell University, Ithaca, NY 14853, USA \\
    $^{2}$Department of Materials Science and Engineering, Cornell University, Ithaca, NY 14853, USA \\
    $^{3}$Center for Applied Mathematics, Cornell University, Ithaca, NY 14853, USA \\
    $^{4}$Kavli Institute at Cornell for Nanoscale Science, Cornell University, Ithaca, NY 14853, USA \\
    $^{5}$School of Operations Research and Information Engineering, Cornell University, Ithaca, NY 14853, USA \\[0.4em]
\texttt{\{%
\href{mailto:dy327@cornell.edu}{dy327},%
\href{mailto:pb482@cornell.edu}{pb482},%
\href{mailto:chia-hao.lee@cornell.edu}{chia-hao.lee},%
\href{mailto:yz685@cornell.edu}{yz685},%
\href{mailto:david.a.muller@cornell.edu}{david.a.muller},%
\href{mailto:pf98@cornell.edu}{pf98}%
\}@cornell.edu}
  \end{tabular}%
}

\begin{document}

\maketitle

\begin{abstract}
%% 20260824 - CHL attempt 9pm. DY endorsed. 
Estimating physical parameters from scientific images is a common inverse problem in materials characterization that often relies on expensive physics-based simulations. In electron microscopy, specimen thickness and crystal mistilt are critical parameters that govern how electrons scatter through the sample, and therefore the accuracy of any atomic-scale structure recovered from it. They are commonly inferred by matching experimental position-averaged convergent-beam electron diffraction (PACBED) patterns to simulated ones, but grid searches scale poorly and neural-network methods require extensive pretraining that may not transfer to new conditions.
Here, we propose scalable Bayesian optimization of composite functions (SBOCF), a simulation-efficient method that exploits the known composite structure of the image-matching objective and the intermediate information contained in simulated images. By representing PACBED images with patch-level summaries and two correction terms, SBOCF preserves the original pixel-wise objective while reducing the number of modeled outputs from $24{,}649$ to $11$. %Under a budget of 50 simulator evaluations, SBOCF achieved a median final SSE $290\times$ lower than \pbedit{standard Bayesian Optimization} with expected improvement on a 380~\textup{\AA} synthetic \ce{SrTiO3} specimen, and produced more accurate parameter estimates on both synthetic benchmarks. On experimental data, its estimates agreed with previously reported values without task-specific pretraining. 
Under a budget of 50 simulator evaluations, SBOCF outperformed standard Bayesian optimization with expected improvement on synthetic SrTiO$_3$ benchmarks with thick and thin specimens, reducing the median final SSE by up to $290\times$ in the thick-sample case. On experimental data, SBOCF produced parameter estimates consistent with previously reported values without task-specific pretraining. For a simulated mistilted specimen, using the SBOCF estimates in a downstream ptychographic reconstruction recovered sharp atoms that were otherwise blurred. These results establish SBOCF as a promising approach for inverse problems involving expensive simulators and high-dimensional structured outputs.

\end{abstract}

\section{Introduction}\label{sec:intro}

Estimating unknown parameters by matching experimental measurements to physics-based simulations is a common inverse problem across scientific fields, where the cost of each simulation can limit characterization throughput. Electron microscopy is a representative instance: a beam of high-energy electrons is transmitted through a thin specimen, scattering off the atoms inside, and a detector records the electrons that emerge. Because the scattering depends on how the atoms are arranged, this signal encodes the atomic structure, but recovering the atomic positions requires a computational reconstruction. Such structural analysis depends on two quantities rarely known beforehand: the specimen \emph{thickness}, which sets how many times electrons scatter; and the crystal \emph{mistilt}, the angular deviation between the beam and the crystallographic zone axis, the direction along which the atoms line up into columns. Exact zone-axis alignment is difficult in practice, so a residual mistilt of a few milliradians is typical, and errors in either quantity propagate into the inferred atomic positions.
%the specimen \emph{thickness}, which sets how far the electrons travel and hence how many times they scatter; and the crystal \emph{mistilt}, the angular deviation between the beam and the crystallographic zone axis, the direction along which the atoms line up into columns. Exact zone-axis alignment is difficult in practice, so a residual mistilt of a few milliradians is typical, and errors in either quantity propagate into the inferred atomic positions.
%Automated materials characterization often requires estimating unknown physical parameters by matching experimental measurements with physics-based simulations. When each simulation is computationally expensive, searching over candidate parameter values can become a major bottleneck, limiting the throughput of characterization workflows. This challenge arises in electron microscopy, where accurate specimen-parameter estimates are essential for obtaining reliable structural information.

Both parameters can be estimated from the same measurement used for structural analysis. In four-dimensional scanning transmission electron microscopy (4D-STEM), a focused electron probe is scanned across the specimen and a two-dimensional diffraction pattern is recorded at each position, giving a four-dimensional dataset. Averaging these patterns over probe positions spanning at least one unit cell yields the position-averaged convergent-beam electron diffraction (PACBED) pattern, a single image whose fringe structure and symmetry depend sensitively on thickness and tilt~\cite{lebeau_position_2010}. The mapping from a PACBED pattern back to the parameters that produced it is not available in closed form, but the forward direction is: for candidate values, a multislice simulation~\cite{kirkland_advanced_2010} propagates the electron wave through the specimen slice by slice and predicts the resulting pattern. Estimating thickness and mistilt is therefore an inverse problem solved by forward simulation: we seek the parameters whose simulated pattern best matches the measured one, and each simulation is expensive. These estimates are valuable on their own and also benefit downstream methods: multislice electron ptychography (MEP)~\cite{chen_electron_2021} recovers an atomic-resolution, depth-resolved image from the same 4D-STEM dataset, but degrades when thickness and mistilt are inaccurate~\cite{sha_deep_2022}; estimating them beforehand removes them from the reconstruction's variables.
%In this work, we consider multislice electron ptychography (MEP), which has received growing interest for atomic-scale three-dimensional characterization since its first experimental demonstration~\cite{chen_electron_2021}. The quality of MEP reconstructions depends strongly on specimen parameters such as thickness and mistilt. Precise alignment to a zone axis can be difficult in experimental settings, making some degree of mistilt unavoidable. Although known tilt angles can be incorporated into the multislice propagator~\cite{kirkland_advanced_2010,sha_deep_2022}, estimating thickness and tilt jointly with the other reconstruction variables can substantially increase computational cost and reconstruction time.

%Because MEP processes four-dimensional scanning transmission electron microscopy (4D-STEM) datasets, the acquired measurements can also be used to estimate specimen parameters before reconstruction. In particular, position-averaged convergent beam electron diffraction (PACBED) patterns are two-dimensional reductions of 4D-STEM data that contain information about specimen thickness and tilt~\cite{lebeau_position_2010}. Given a candidate set of specimen parameters, a multislice simulator~\cite{kirkland_advanced_2010} can generate the corresponding PACBED pattern. PACBED-based parameter estimation can therefore be formulated as an inverse problem: given an experimental PACBED image, we seek the specimen parameters whose simulated PACBED image best matches the measurement.

Existing PACBED-based approaches face a trade-off between cost and generality. Conventional methods perform a grid search, exhaustively simulating patterns over a dense grid of thickness and tilt values and comparing them via least-squares metrics ~\cite{pollock_accuracy_2017}. This scales poorly as the parameter space grows. More recent methods train convolutional neural networks to estimate parameters rapidly ~\cite{xu_deep_2018,zhang_atomic_2020,oberaigner_online_2023}, but they are tied to specific material systems and imaging conditions, requiring large simulated libraries and retraining for new settings.

Bayesian optimization (BO)~\cite{movckus1975bayesian,frazier_tutorial_2018} reduces costly simulations by using a surrogate model of the final image-discrepancy objective to guide optimization, without requiring a task-specific training dataset. However, this scalar model does not exploit the known way pixel intensities are combined to compute the discrepancy. Bayesian optimization of composite functions (BOCF)~\cite{astudillo_bayesian_2019} exploits this structure by modeling individual pixel intensities, but the number of model outputs makes it computationally challenging for high-resolution images with larger number of pixels.

To address this challenge, we introduce \emph{scalable Bayesian optimization of composite functions} (SBOCF). SBOCF partitions each image into patches and expresses the original image-discrepancy objective in a new composite form defined over lower-dimensional patch summaries and correction terms. This reduces the number of modeled outputs by more than 2,000-fold, retaining BOCF’s ability to exploit the objective’s composite structure while making it practical for high-resolution PACBED images. We validate SBOCF on synthetic and experimental SrTiO$_3$ PACBED benchmarks. It yields more accurate parameter estimates than standard BO and random sampling on synthetic data, and estimates consistent with earlier neural-network-based estimates on experimental data, without requiring a large task-specific training set. We also demonstrate that SBOCF estimates improve downstream MEP quality. More broadly, SBOCF offers a promising approach for inverse problems involving expensive simulators and high-dimensional structured outputs.

\begin{figure}[htp]
    \centering

    \includegraphics[width=0.7\textwidth]
    {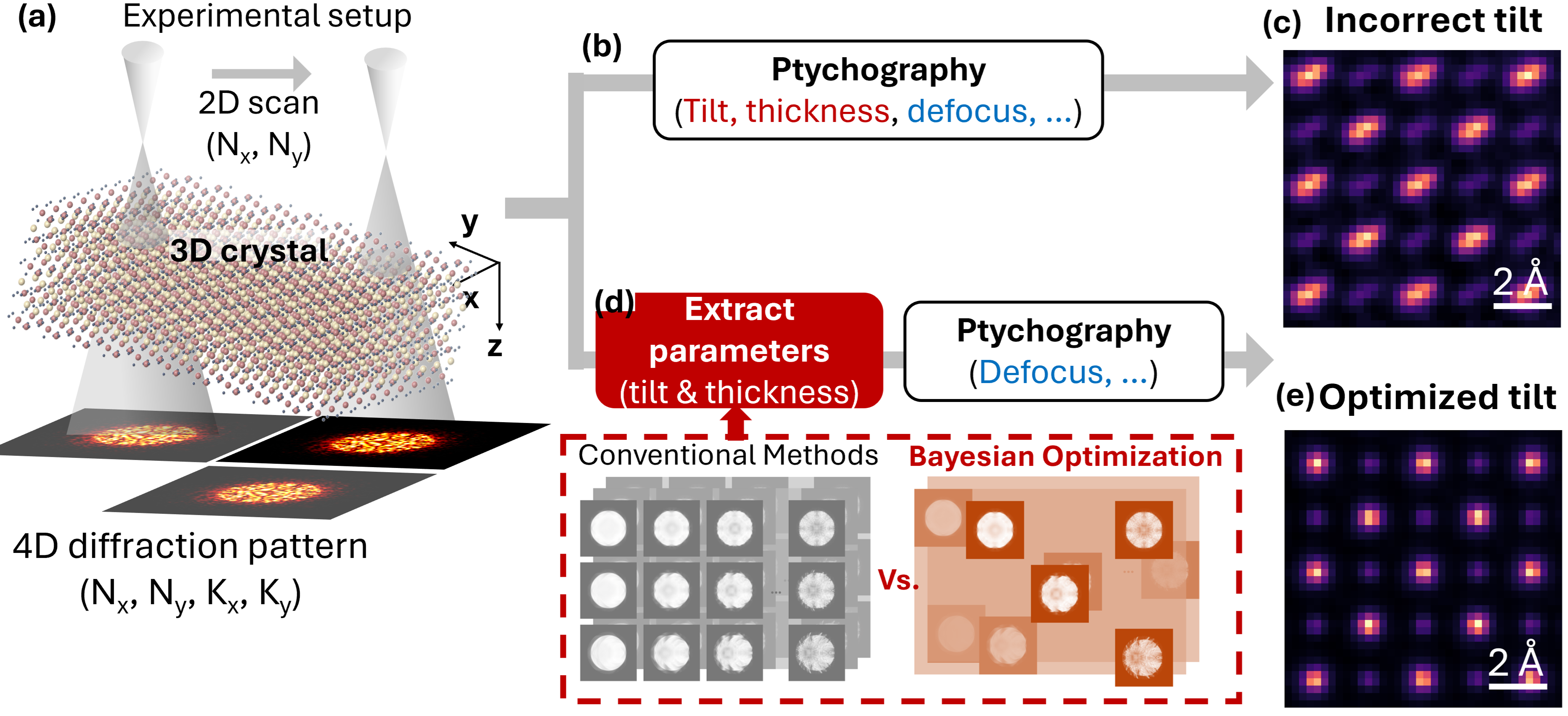}

    % \vspace{0.5em}

    % \resizebox{\textwidth}{!}{%
    % \input{tikz_code_for_SBOCF_loop.tex}%
    % }

    \caption{
    Conceptual workflow for PACBED-based specimen-parameter estimation prior to MEP. (a) A 4D-STEM dataset is acquired from a specimen with unknown thickness and mistilt. (b) Conventionally these parameters are estimated jointly with the other reconstruction variables, and (c) uncorrected mistilt degrades the result. (d) In the proposed workflow they are estimated first, by comparing the experimental PACBED pattern with multislice simulations: conventional methods evaluate a predefined grid, whereas BO-based methods adaptively select candidates from previous results. (e) The improved reconstruction expected after correction.
    }
    \label{fig:workflow}
\end{figure}

Fig.~\ref{fig:workflow} illustrates the complete workflow, from experimental 4D-STEM acquisition to PACBED-based specimen-parameter estimation using BO-based methods.

%We evaluate SBOCF against standard BO and random sampling on synthetic SrTiO$_3$ PACBED benchmarks with known specimen parameters and on an experimental SrTiO$_3$ PACBED benchmark. On the synthetic problems, SBOCF achieves lower pixel-wise SSE and more accurate parameter estimates than the competing methods. On the experimental benchmark, SBOCF obtains parameter estimates consistent with previously reported values without requiring a large task-specific training dataset. These results demonstrate the potential of SBOCF to reduce the simulation burden associated with PACBED-based parameter estimation and support more efficient automated materials-characterization workflows.
% \begin{figure}[htp]
%     \centering
%     \includegraphics[width=0.8\textwidth]{figures/main-fig1-workflow.png}
%     \vspace{0.5em}

%     \resizebox{0.95\textwidth}{!}{%
%         \input{figures/sbocf-flowchart.tex}%
%         }
%     \caption{
% Conceptual workflow for PACBED-based specimen-parameter estimation prior to multislice electron ptychography. (a) A 4D-STEM dataset is acquired by scanning an electron probe across a specimen with unknown thickness and mistilt. (b) In a conventional workflow, these specimen parameters may be estimated jointly with other ptychographic reconstruction variables. (c) Illustration of a reconstruction degraded by uncorrected mistilt. (d) In the proposed workflow, thickness and tilt are estimated before reconstruction by comparing the experimental PACBED pattern with patterns generated by multislice simulations. Conventional methods evaluate a predefined grid of candidate parameters, whereas BO-based methods adaptively select candidates using previous simulation results. (e) Illustration of the improved reconstruction expected after correcting the specimen tilt. (f) Flowchart of the proposed SBOCF algorithm.
% }
%     \label{fig:workflow}
% \end{figure}
\section{PACBED Parameter Estimation with Bayesian Optimization}
\label{sec:bo}

\subsection{Problem formulation}
\label{sec:problem}

We consider a target PACBED image with $M\times M$ pixels. Let
$\mathbf{y}^{\mathrm{true}}\in\mathbb{R}^{M\times M}$ denote the experimental PACBED image, where $y^{\mathrm{true}}_{k,\ell}$ is the intensity at pixel $(k,\ell)$. Let $\mathcal{X}\subseteq\mathbb{R}^d$ denote the specimen parameter space. In this work, we consider $d=3$ parameters: specimen thickness and the tilt components along the $x$- and $y$-directions.

For a given parameter vector $\mathbf{x}\in\mathcal{X}$, a multislice simulation generates a PACBED image
$\mathbf{y}(\mathbf{x})\in\mathbb{R}^{M\times M}$, where $y_{k,\ell}(\mathbf{x})$ denotes the simulated intensity at pixel $(k,\ell)$. We measure the discrepancy between the simulated and experimental PACBED patterns using the pixel-wise sum of squared errors (SSE),
\begin{equation}
    f(\mathbf{x})
    :=
    \sum_{k=1}^{M}\sum_{\ell=1}^{M}
    \left(
    y_{k,\ell}(\mathbf{x})
    -
    y^{\mathrm{true}}_{k,\ell}
    \right)^2.
    \label{eq:sse}
\end{equation}
The parameter estimation problem is then
\begin{equation}
    \mathbf{x}^{*}
    \in
    \arg\min_{\mathbf{x}\in\mathcal{X}}
    f(\mathbf{x}).
    \label{eq:optimization_problem}
\end{equation}

\begin{figure}
    \centering
    \includegraphics[width=0.68\linewidth]{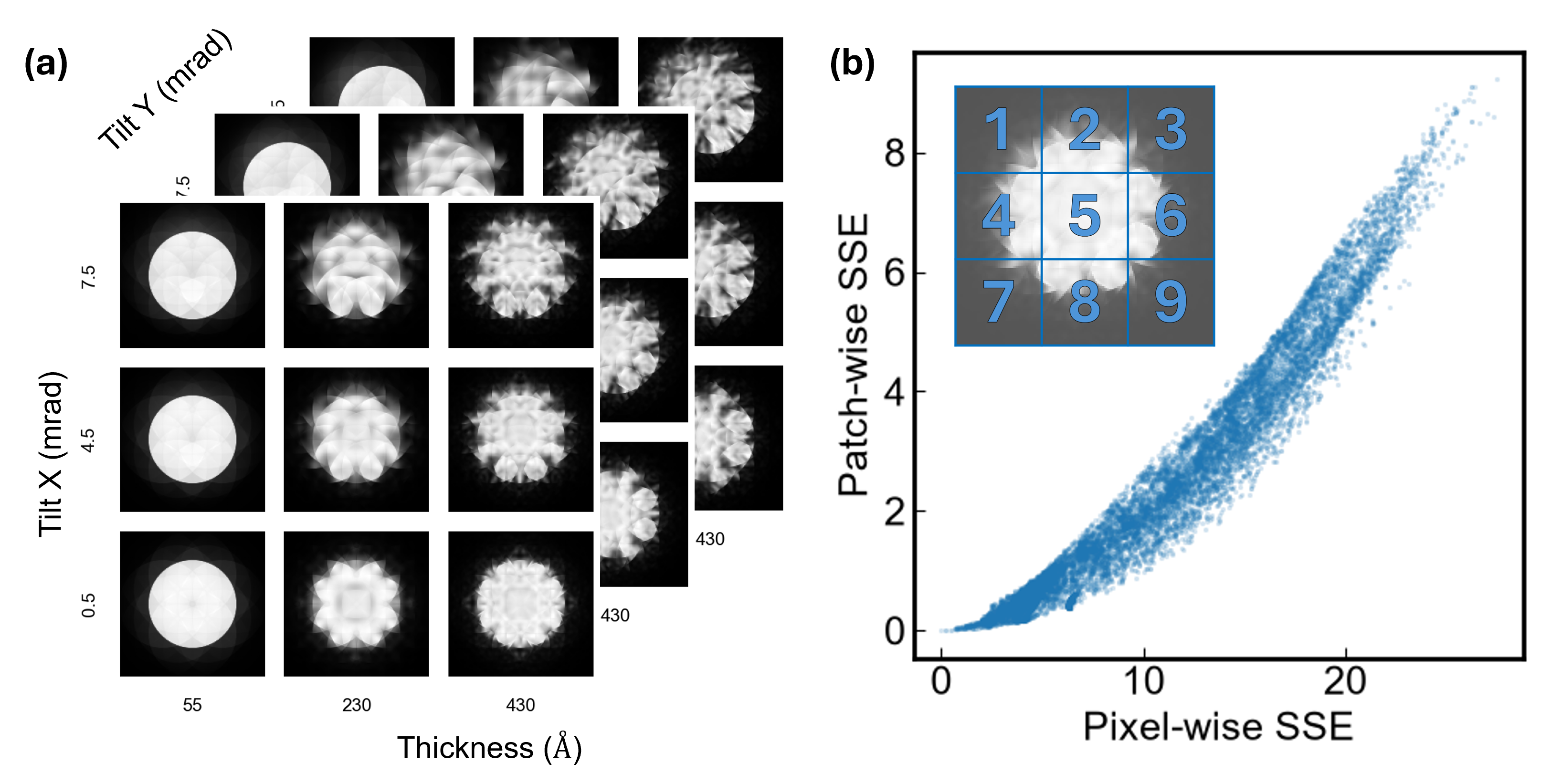}
    \caption{(a) PACBED patterns simulated for SrTiO$_3$ across thickness (columns) and tilt along two orthogonal axes, tiltX (rows) and tiltY (offset diagonally). Thickness reorganizes the fringe structure non-monotonically, while tilt along either axis breaks the pattern's symmetry in a direction-dependent way. (b) Pixel-wise SSE $f(\mathbf{x})$ versus patch-level SSE $f^{P}(\mathbf{x})$ for simulated PACBED images; their systematic relationship motivates using the patch-level SSE with correction terms.}
    \label{fig:patch_pixel_sse}
\end{figure}

Fig.~\ref{fig:patch_pixel_sse}(a) illustrates how the simulated pattern $\by(\bx)$ varies over the thickness and tilt ranges considered in this work.

\subsection{Bayesian optimization}
\label{sec:standard_bo}

Bayesian optimization (BO) is a sequential approach for optimizing expensive black-box functions, applied to problems such as hyperparameter tuning~\cite{snoek2012practical}, materials design~\cite{frazier2015bayesian}, and experimental optimization~\cite{khongkomolsakul2025improving}. In electron microscopy it has been used to select reconstruction parameters for electron ptychography~\cite{cao2022automatic} and to align the microscope~\cite{pratiush2026towards}. Both target the instrument or the reconstruction; we instead use BO to estimate specimen thickness and mistilt by matching simulated and experimental PACBED patterns, then supply the estimates to a subsequent ptychographic reconstruction.

Standard BO places a Gaussian process (GP) prior on the scalar objective \(f(\mathbf{x})\). Given observations $
\mathcal{D}_n
=
\left\{
\left(\mathbf{x}_i,f(\mathbf{x}_i)\right)
\right\}_{i=1}^{n},
$
the GP posterior yields the predictive distribution
\[
f(\mathbf{x})\mid\mathcal{D}_n
\sim
\mathcal{N}\!\left(
\mu_n(\mathbf{x}),
\sigma_n^2(\mathbf{x})
\right),
\qquad \mathbf{x}\in\mathcal{X},
\]
where \(\mu_n(\mathbf{x})\) and \(\sigma_n^2(\mathbf{x})\) are the posterior mean and variance, respectively; see Appendix~\ref{app:gpformulas}. An acquisition function uses this distribution to balance exploitation of inputs with promising predicted objective values and exploration of inputs with high posterior uncertainty.

A widely used acquisition function for BO is expected improvement (EI)~\cite{jones_efficient_1998},
\begin{equation}
\alpha_n^{\mathrm{EI}}(\mathbf{x})
=
\mathbb{E}_n
\left[
\max\left\{
f_n^{\min}-f(\mathbf{x}),0
\right\}
\right],
\label{eq:ei}
\end{equation}
where $f_n^{\min}=\min_{i=1,\ldots,n}f(\mathbf{x}_i)$ is the best objective value observed so far. EI measures the expected improvement over this value and admits a closed-form expression under a Gaussian predictive distribution (Appendix~\ref{app:eiformula}). At each iteration, BO with EI selects
\begin{equation}
\mathbf{x}_{n+1}
\in
\arg\max_{\mathbf{x}\in\mathcal{X}}
\alpha_n^{\mathrm{EI}}(\mathbf{x}).
\label{eqn:ei_opt}
\end{equation}
A PACBED pattern is then simulated at \(\mathbf{x}_{n+1}\), its SSE \(f(\mathbf{x}_{n+1})\) is evaluated, and the observation is added to \(\mathcal{D}_n\) to form \(\mathcal{D}_{n+1}\). This process repeats until the evaluation budget is exhausted. 
% We compare our proposed method against BO using EI and BO using the knowledge gradient (KG)~\cite{frazier_knowledge-gradient_2008,scott2011correlated}, defined in Appendix~\ref{app:KGformula}.}

% We primarily use the expected improvement (EI) acquisition function~\cite{jones_efficient_1998},
% \begin{equation}
% \alpha_n^{\mathrm{EI}}(\mathbf{x})
% =
% \mathbb{E}_n
% \left[
% \max\left\{
% f_n^{\min}-f(\mathbf{x}),0
% \right\}
% \right],
% \label{eq:ei}
% \end{equation}
% where $f_n^{\min}=\min_{i=1,\ldots,n}f(\mathbf{x}_i)$
% is the best objective value observed so far. Under the Gaussian predictive distribution of $f(\bx)$, EI admits a closed-form expression given in Appendix~\ref{app:eiformula}. At each iteration, BO selects
% \begin{equation}
% \mathbf{x}_{n+1}
% \in
% \arg\max_{\mathbf{x}\in\mathcal{X}}
% \alpha_n^{\mathrm{EI}}(\mathbf{x}).
% \label{eqn:ei_opt}
% \end{equation}
% A PACBED pattern is then simulated at \(\mathbf{x}_{n+1}\), its SSE \(f(\mathbf{x}_{n+1})\) is evaluated, and the resulting observation is added to \(\mathcal{D}_n\) to form \(\mathcal{D}_{n+1}\). This process is repeated until the evaluation budget is exhausted. We also consider the knowledge gradient (KG)~\cite{frazier_knowledge-gradient_2008,scott2011correlated} as an additional BO baseline; its definition is provided in Appendix~\ref{app:KGformula}.

While standard BO models $f(\mathbf{x})$ directly as a scalar black-box objective, the PACBED objective in Eq.~\eqref{eq:sse} has additional known structure: the specimen parameters $\bx$ are first mapped through the simulator to a PACBED image $\by(\bx)$, and the resulting pixel intensities are then combined through the SSE function. We present how to exploit this structure to improve BO sampling efficiency in the next section.

\subsection{Bayesian optimization of composite functions}
\label{sec:bocf}

The PACBED objective in Eq.~\eqref{eq:sse} has the known composite form
\begin{equation}
f(\mathbf{x})
=
g\!\left(\mathbf{y}(\mathbf{x})\right),
\label{eq:composite}
\end{equation}
where the known outer function \(g\) takes the form:
$
g(\mathbf{y})
=
\sum_{k=1}^{M}
\sum_{\ell=1}^{M}
\left(
y_{k,\ell}
-
y^{\mathrm{true}}_{k,\ell}
\right)^2.
\label{eq:bocf_outer}
$

Bayesian optimization of composite functions (BOCF)~\cite{astudillo_bayesian_2019} exploits this structure by modeling the intermediate output \(\mathbf{y}(\mathbf{x})\) with a multi-output GP and propagating its predictive uncertainty through \(g\) to obtain a predictive distribution over \(f(\mathbf{x})\). By exploiting this additional structure, BOCF has been shown to reduce simple regret by several orders of magnitude relative to standard BO on benchmark problems~\cite{astudillo_bayesian_2019}, motivating its use for simulation-intensive inverse problems. The resulting predictive distribution is used to evaluate EI and select the next candidate. Because nonlinear \(g\) generally induces a non-Gaussian distribution over \(f(\mathbf{x})\), EI typically lacks a closed-form expression and is estimated using Monte Carlo sampling~\cite{astudillo_bayesian_2019}.
% Bayesian optimization of composite functions (BOCF)~\cite{astudillo_bayesian_2019} exploits this structure by modeling the intermediate output \(\mathbf{y}(\mathbf{x})\) with a multi-output GP and propagating its predictive uncertainty through \(g\) to obtain a predictive distribution over \(f(\mathbf{x})\). This distribution is then used to evaluate EI and select the next candidate. Because applying the nonlinear function \(g\) to the Gaussian predictive distribution of \(\mathbf{y}(\mathbf{x})\) generally produces a non-Gaussian distribution over \(f(\mathbf{x})\), EI in this case does not admit a closed-form expression and is instead estimated using Monte Carlo sampling~\cite{astudillo_bayesian_2019}.

Directly applying standard BOCF to PACBED requires modeling $M^2$ intermediate pixel intensities $\by(\bx)$, which becomes computationally impractical for typical PACBED images with $M \gtrsim 128$. This computational challenge motivates our proposed method, presented in the next section, which uses a lower-dimensional intermediate representation.

\section{Proposed Method: Scalable Bayesian Optimization of Composite Functions}
\label{sec:sbocf}
To address the computational challenge of GP modeling in standard BOCF, we propose \emph{scalable Bayesian optimization of composite functions} (SBOCF). SBOCF replaces the pixel-level intermediate output with lower-dimensional patch-level representations while retaining the original pixel-wise SSE as the optimization objective.

\paragraph{Patch-level representation.}
We partition an $M\times M$ PACBED image into $P$ disjoint patches $
\mathcal{P}=
\left\{
\mathcal{P}_1,\ldots,\mathcal{P}_P
\right\}$.
For each patch $\mathcal{P}_j$, we define a scalar summary
\[
h_j(\mathbf{x})
=
A_j\!\left(\mathbf{y}(\mathbf{x})\right),
\qquad j=1,\ldots,P,
\]
where $A_j$ is an aggregation operator applied to the pixels in $\mathcal{P}_j$. In our experiments, $A_j$ sums the pixel intensities within each patch:
\begin{equation}
A_j\!\left(\mathbf{y}(\mathbf{x})\right)
=
\sum_{(k,\ell)\in\mathcal{P}_j}
y_{k,\ell}(\mathbf{x}).
\label{eqn:sumpixel}
\end{equation}
We similarly define the corresponding target summary as $
h_j^{\mathrm{true}}
=
A_j\!\left(\mathbf{y}^{\mathrm{true}}\right)$.
Collecting the patch summaries gives $
\mathbf{h}(\mathbf{x})
=
\left(
h_1(\mathbf{x}),\ldots,h_P(\mathbf{x})
\right)^\top$.

Using these summaries, we define the patch-level SSE
\begin{equation*}
f^{P}(\mathbf{x})
=
g^{P}\!\left(\mathbf{h}(\mathbf{x})\right)
=
\sum_{j=1}^{P}
\left(
h_j(\mathbf{x})-h_j^{\mathrm{true}}
\right)^2.
\end{equation*}
% This representation reduces the number of intermediate quantities from $M^2$ pixel intensities to $P$ patch summaries. However, aggregation discards spatial information within each patch, so the minimizers of $f^{P}(\mathbf{x})$ are generally not equal to those of the original pixel-wise objective $f(\mathbf{x})$.
This representation reduces the number of intermediate quantities from $M^2$ pixel intensities to $P$ patch summaries. By capturing discrepancies at a coarser spatial scale, $f^{P}(\mathbf{x})$ provides an approximation to the original pixel-wise objective, and its minimizers may therefore approximate those of $f(\mathbf{x})$. However, aggregation discards spatial information within each patch, so the two objectives generally need not share the same minimizers.

\paragraph{Correction for aggregation error.}
% Although $f^{P}(\mathbf{x})$ does not exactly recover $f(\mathbf{x})$, 
Although the minimizers of $f^{P}(\mathbf{x})$ need not coincide with those of $f(\mathbf{x})$, the two quantities exhibit a strong systematic relationship across simulated PACBED images, as shown in Fig.~\ref{fig:patch_pixel_sse}(b). We therefore represent the original objective as
\begin{equation}
f(\mathbf{x})
=
\epsilon(\mathbf{x})f^{P}(\mathbf{x})
+
\delta(\mathbf{x}),
\label{eq:sbocf_decomposition}
\end{equation}
where $\epsilon(\mathbf{x})$ is an input-dependent multiplicative correction and $\delta(\mathbf{x})$ captures the remaining discrepancy. This decomposition is not unique. We therefore constrain the multiplicative correction to a prescribed range and select the decomposition with the smallest absolute residual:
\begin{equation}
\begin{aligned}
\min_{\epsilon(\mathbf{x}),\,\delta(\mathbf{x})}
&\quad
|\delta(\mathbf{x})|
\\
\mathrm{s.t.}
&\quad
|\log_{10}\epsilon(\mathbf{x})|
\leq c,
\quad\text{ and }
\quad
f(\mathbf{x})
=
\epsilon(\mathbf{x})f^{P}(\mathbf{x})
+
\delta(\mathbf{x}),
\end{aligned}
\label{eq:sbocf_decomposition_opt}
\end{equation}
where $c>0$ controls the allowable range of the multiplicative correction. This formulation assigns the systematic difference between $f^{P}(\mathbf{x})$ and $f(\mathbf{x})$ primarily to the bounded scaling term, while $\delta(\mathbf{x})$ captures the discrepancy that cannot be explained by this scaling. A closed-form solution of optimization problem \eqref{eq:sbocf_decomposition_opt} is provided in Appendix~\ref{app:eps_del_decom}.

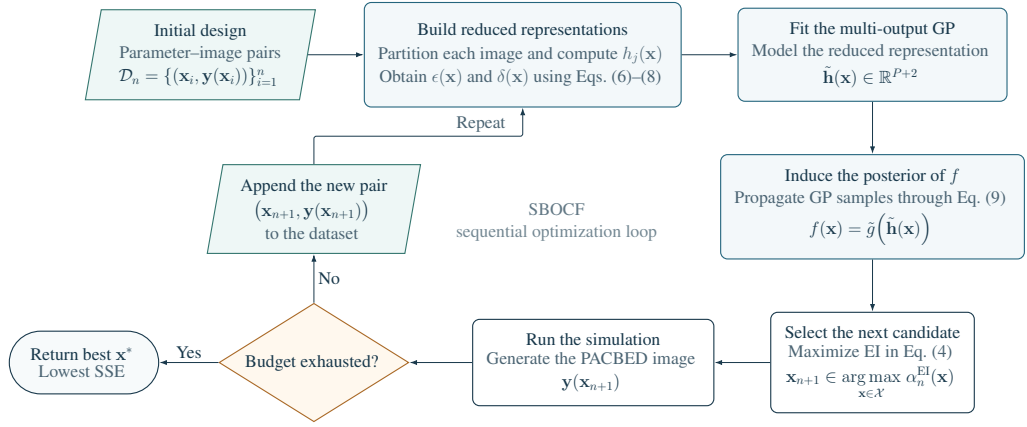
\begin{figure}[h]
    \centering

    \resizebox{\textwidth}{!}{%
        % figures/sbocf-flowchart.tex
%
% Required in the main document preamble:
%
% \usepackage{tikz}
% \usepackage{amsmath,amssymb,bm}
% \usetikzlibrary{
%     arrows.meta,
%     calc,
%     positioning,
%     shapes.geometric,
%     shapes.misc
% }

\begingroup

%------------------------------------------------
% Colors
%------------------------------------------------
\definecolor{flowNavy}{HTML}{18384B}
\definecolor{flowTeal}{HTML}{24657D}
\definecolor{flowGreen}{HTML}{34766F}
\definecolor{flowMethodFill}{HTML}{F3F8FA}
\definecolor{flowDataFill}{HTML}{EEF7F6}
\definecolor{flowDecisionFill}{HTML}{FFF8EE}
\definecolor{flowDecisionLine}{HTML}{A96825}
\definecolor{flowTermFill}{HTML}{F7FAFB}

%------------------------------------------------
% Text styles
%------------------------------------------------
\newcommand{\flowheading}{%
    \normalfont\rmfamily\Large\color{flowNavy}%
}

\newcommand{\flowbody}{%
    \normalfont\rmfamily\Large\color{flowNavy!78}%
}

\newcommand{\flowequation}{%
    \normalfont\rmfamily\Large\color{flowNavy}%
}

%------------------------------------------------
% TikZ styles
%------------------------------------------------
\tikzset{
    flowarrow/.style={
        -{Latex[length=2.2mm,width=1.5mm]},
        draw=flowNavy,
        line width=0.8pt,
        rounded corners=2pt
    },
    process/.style={
        rectangle,
        rounded corners=2mm,
        draw=flowNavy,
        fill=white,
        line width=0.9pt,
        align=center,
        inner sep=4mm
    },
    method/.style={
        process,
        draw=flowTeal,
        fill=flowMethodFill
    },
    data/.style={
        trapezium,
        trapezium left angle=80,
        trapezium right angle=100,
        draw=flowGreen,
        fill=flowDataFill,
        line width=0.9pt,
        align=center,
        inner xsep=5mm,
        inner ysep=4mm
    },
    decision/.style={
        diamond,
        aspect=1.6,
        draw=flowDecisionLine,
        fill=flowDecisionFill,
        line width=0.9pt,
        align=center,
        inner sep=1.5mm
    },
    terminator/.style={
        rounded rectangle,
        rounded rectangle arc length=180,
        draw=flowNavy,
        fill=flowTermFill,
        line width=0.9pt,
        align=center,
        inner xsep=5mm,
        inner ysep=3mm
    },
    branch/.style={
        font=\normalfont\rmfamily\Large,
        text=flowNavy
    },
    annotation/.style={
        font=\normalfont\rmfamily\Large,
        text=flowNavy!78
    }
}

%------------------------------------------------
% Flowchart
%------------------------------------------------
\begin{tikzpicture}[
    font=\normalfont\rmfamily,
    node distance=12mm and 16mm
]

%------------------------------------------------
% Top row
%------------------------------------------------

\node[
    data,
    minimum width=38mm,
    minimum height=24mm
] (initial) {%
    {\flowheading Initial design}\\[1.5mm]
    {\flowbody Parameter--image pairs}\\[1.5mm]
    {\flowequation
        $\mathcal D_n
        =
        \{(\mathbf{x}_i,\mathbf{y}(\mathbf{x}_i))\}_{i=1}^{n}$
    }
};

\node[
    method,
    right=of initial,
    minimum width=65mm,
    minimum height=30mm
] (representation) {%
    {\flowheading Build reduced representations}\\[1.5mm]
    {\flowbody
        Partition each image and compute $h_j(\mathbf{x})$
    }\\[1mm]
    {\flowbody
        Obtain $\epsilon(\mathbf{x})$ and $\delta(\mathbf{x})$
        using Eqs.~\eqref{eqn:sumpixel}--%
        \eqref{eq:sbocf_decomposition_opt}
    }
};

\node[
    method,
    right=of representation,
    minimum width=46mm,
    minimum height=24mm
] (gp) {%
    {\flowheading Fit independent GP surrogates}\\[1.5mm]
    {\flowbody Model the reduced representation}\\[1.5mm]
    {\flowequation
        $\tilde{\mathbf{h}}(\mathbf{x})
        \in \mathbb R^{P+2}$
    }
};

%------------------------------------------------
% Right column
%------------------------------------------------

\node[
    method,
    below=15mm of gp,
    minimum width=46mm,
    minimum height=22mm
] (posterior) {%
    {\flowheading Induce the posterior of $f$}\\[1.5mm]
    {\flowbody
        Propagate GP samples through
        Eq.~\eqref{eq:sbocf_composite}
    }\\[1.5mm]
    {\flowequation
        $f(\mathbf{x})
        =
        \tilde{g}\!\left(
            \tilde{\mathbf{h}}(\mathbf{x})
        \right)$
    }
};

\node[
    process,
    below=15mm of posterior,
    minimum width=46mm,
    minimum height=22mm
] (select) {%
    {\flowheading Select the next candidate}\\[1.5mm]
    {\flowbody
        Maximize EI in Eq.~\eqref{eqn:ei_opt}
    }\\[1.5mm]
    {\flowequation
        $\mathbf{x}_{n+1}
        \in
        \underset{\mathbf{x}\in\mathcal X}
        {\operatorname{arg\,max}}
        \;
        \alpha_n^{\mathrm{EI}}(\mathbf{x})$
    }
};

%------------------------------------------------
% Bottom row and dataset update
%------------------------------------------------

\node[
    process,
    left=of select,
    minimum width=43mm,
    minimum height=20mm
] (simulate) {%
    {\flowheading Run the simulation}\\[1.5mm]
    {\flowbody Generate the PACBED image}\\[1.5mm]
    {\flowequation $\mathbf{y}(\mathbf{x}_{n+1})$}
};

\node[
    data,
    left=18mm of simulate,
    minimum width=42mm,
    minimum height=19mm
] (append) {%
    {\flowheading Append the new pair}\\[1.5mm]
    {\flowequation
        $\bigl(
            \mathbf{x}_{n+1},
            \mathbf{y}(\mathbf{x}_{n+1})
        \bigr)$
    }\\[1.2mm]
    {\flowbody to the dataset}\\[1.2mm]
    {\flowequation $n\leftarrow n+1$}
};

\node[
    decision,
    above=14mm of append,
    minimum width=27mm,
    minimum height=22mm
] (budget) {%
    {\flowheading Budget exhausted?}
};

\node[
    terminator,
    left=17mm of budget,
    minimum width=28mm,
    minimum height=18mm
] (return) {%
    {\flowheading Return best $\mathbf{x}^{*}$}\\[1.2mm]
    {\flowbody Lowest observed SSE}
};

%------------------------------------------------
% Main arrows
%------------------------------------------------

\draw[flowarrow]
    (initial.east) -- (representation.west);

\draw[flowarrow]
    (representation.east) -- (gp.west);

\draw[flowarrow]
    (gp.south) -- (posterior.north);

\draw[flowarrow]
    (posterior.south) -- (select.north);

\draw[flowarrow]
    (select.west) -- (simulate.east);

\draw[flowarrow]
    (simulate.west) -- (append.east);

\draw[flowarrow]
    (append.north) -- (budget.south);

%------------------------------------------------
% Decision branches and sequential update loop
%------------------------------------------------

\draw[flowarrow]
    (budget.west)
    --
    node[branch, above] {Yes}
    (return.east);

% \draw[flowarrow]
%     (budget.north)
%     --
%     ++(0,8mm)
%     -|
%     node[pos=0.40, below, branch] {No: repeat}
%     (representation.south);
\draw[flowarrow]
    (budget.north)
    |-
    node[pos=0.75, below=1mm, branch] {No: repeat}
    ([yshift=-8mm]representation.south)
    --
    (representation.south);
%------------------------------------------------
% Center label
%------------------------------------------------

\node[
    font=\normalfont\rmfamily\Large,
    text=flowNavy!68
] (centerlabel)
at ($(representation.south)!0.48!(simulate.north)$) {
    SBOCF
};

\node[
    font=\normalfont\rmfamily\Large,
    text=flowNavy!60,
    below=0.3mm of centerlabel
] {
    sequential optimization loop
};

\end{tikzpicture}

\endgroup%
    }

    \caption{
        Flowchart of the proposed SBOCF algorithm.
    }
    \label{fig:sbocf-flowchart}
\end{figure}

\paragraph{Composite surrogate model.}
SBOCF models the reduced intermediate representation
$
\tilde{\mathbf{h}}(\mathbf{x})
=
\left(
h_1(\mathbf{x}),
\ldots,
h_P(\mathbf{x}),
\epsilon(\mathbf{x}),
\delta(\mathbf{x})
\right)^\top
$
using independent GP surrogates for its $P+2$ outputs. It then applies BOCF to this reduced representation rather than to the full pixel-level PACBED image. The original pixel-wise SSE is recovered through the known outer function
\begin{equation}
f(\mathbf{x})
=
\tilde{g}\!\left(
\tilde{\mathbf{h}}(\mathbf{x})
\right),
\label{eq:sbocf_composite}
\end{equation}
where $\tilde{g}$ takes the form:
$
    \tilde{g}(\tilde{\bh})=\tilde{h}_{P+1}
\sum_{j=1}^{P}
\left(
\tilde{h}_j-h_j^{\mathrm{true}}
\right)^2
+
\tilde{h}_{P+2}.
$
Thus, SBOCF reduces the number of modeled intermediate outputs from $M^2$ to $P+2$ while preserving the original pixel-wise SSE objective.
\paragraph{Sequential SBOCF procedure.}
The complete SBOCF procedure is summarized in
Fig.~\ref{fig:sbocf-flowchart}. Starting from an initial set of parameter--image pairs, SBOCF constructs the reduced representations, fits independent GP surrogates for the $P+2$ outputs, and propagates
posterior samples through the known outer function. It then maximizes EI to select and simulate the next candidate, appends the resulting parameter--image pair to the dataset, and repeats this process until the simulation budget is exhausted.
% \begin{figure}[htp]
%     \centering
%     \includestandalone[
%         width=\textwidth
%     ]{tikz_code_for_SBOCF_loop.tex}
%     \caption{
%         Flowchart of the proposed SBOCF algorithm. Starting from an initial
%         set of parameter--image pairs, SBOCF constructs reduced image
%         representations, fits a multi-output Gaussian process, propagates
%         posterior samples through the known composite objective, and
%         maximizes expected improvement to select the next candidate. This
%         sequential procedure continues until the simulation budget is
%         exhausted.
%     }
%     \label{fig:sbocf-flowchart}
% \end{figure}
\paragraph{Comparison of methods.}

Table~\ref{tab:BOvsSBOCF}  compares the image processing, GP outputs, and objective representations of standard BO, BOCF, and SBOCF. We use the $3\times3$ square partition shown in Fig.~\ref{fig:patch_pixel_sse}(b) throughout the main experiments.
% Table~\ref{tab:BOvsSBOCF} summarizes the differences among standard BO, BOCF, and SBOCF. Row~(a) describes how each method processes the simulated PACBED image before GP modeling: standard BO reduces the image to a scalar SSE, BOCF retains all pixel intensities, and SBOCF partitions the image into patches. We use the $3\times3$ square partition shown in the table throughout the main experiments. Row~(b) lists the quantities modeled by the GP surrogates, while Row~(c) gives the corresponding numbers of GP outputs: $1$ for standard BO, $M^2$ for BOCF, and $P+2$ for SBOCF. Finally, Row~(d) compares their objective representations: standard BO models $f(\mathbf{x})$ directly, BOCF uses the full composite form $g(\mathbf{y}(\mathbf{x}))$, and SBOCF uses the reduced composite form $\tilde{g}(\tilde{\mathbf{h}}(\mathbf{x}))$.
\begin{table*}[h!]
\centering
\caption{Comparison of standard BO, BOCF, and SBOCF for PACBED parameter estimation.}
\label{tab:BOvsSBOCF}

\renewcommand{\arraystretch}{1.35}
\setlength{\tabcolsep}{6pt}

\begin{tabular}{
    p{0.18\textwidth}
    p{0.18\textwidth}
    p{0.22\textwidth}
    p{0.32\textwidth}
}
\toprule

\textbf{Aspect}
&
\centering\textbf{BO}
&
\centering\textbf{BOCF}
&
\centering\arraybackslash\textbf{SBOCF (Ours)}
\\

\midrule

\textbf{(a) Image processing before GP modeling}
&
\centering

Compute scalar SSE
&
\centering
Use pixel intensities directly
&
\centering\arraybackslash

Partition image into patches

\vspace{0.4em}

\\

\midrule

\textbf{(b) GP modeling unit}
&
\centering
Scalar objective

$f(\mathbf{x})$
&
\centering
Pixel intensities

$y_{k,\ell}(\mathbf{x})$
&
\centering\arraybackslash
Patch-level summaries + corrections

$h_j(\mathbf{x}),\epsilon(\mathbf{x}),\delta(\mathbf{x})$
\\

\midrule

\textbf{(c) Number of GP outputs}
&
\centering
$1$

(scalar objective)
&
\centering
$M^2$

(all pixel intensities)
&
\centering\arraybackslash
$P+2$

($P$ patch summaries $+\ \epsilon,\delta$)
\\

\midrule

\textbf{(d) Objective representation}
&
\centering
Direct (black-box)

$f(\mathbf{x})$
&
\centering
Composite

$g(\mathbf{y}(\bx))$ in Eq.~\eqref{eq:composite}
&
\centering\arraybackslash
Reduced composite

$\tilde{g}\!\left(\tilde{\mathbf{h}}(\mathbf{x})\right)$ in Eq.~\eqref{eq:sbocf_composite}
\\

\bottomrule
\end{tabular}
\end{table*}

\section{Numerical Experiments}
\label{sec:exper}
In this section, we first present the comparison methods, optimization settings, and evaluation metrics, followed by the simulated and experimental PACBED benchmarks used to evaluate SBOCF.

\subsection{Comparison methods and experimental setup}
\label{subsec:setting}

% We compare SBOCF with standard BO using a single-output GP with either expected improvement (BO(EI)) or knowledge gradient (BO(KG)), as well as with random sampling (Random). 
%peter's comment: state that BOCF-SP is additional baseline as ablation
% \pbedit{We compare SBOCF with standard BO using a single-output GP with 
% expected improvement (BO(EI)) and knowledge gradient (BO(KG))\pbedit{--KG is defined in Appendix~\ref{app:KGformula}}, partial scalable Bayesian optimization of composite functions (BOCF-SP), and random sampling (Random). BOCF-SP uses the same objective decomposition and correction rule as SBOCF, but places independent GP surrogates on
% \(\epsilon(x)\), \(f^{P}(x)\), and \(\delta(x)\). It retains the
% multiplicative and additive correction structure while treating the
% patch-level SSE as a single unknown function. This baseline assesses
% the additional benefit of explicitly modeling the patch summaries
% underlying \(f^{P}(x)\).}
We compare SBOCF with standard BO using a single-output GP with expected improvement (BO(EI)) or knowledge gradient (BO(KG)), and with random sampling (Random). KG is defined in Appendix~\ref{app:KGformula}. We also include BOCF with scalar patch SSE (BOCF-SP) as an ablation to assess whether SBOCF's gains arise from explicitly modeling individual patch summaries rather than merely reparameterizing the objective. BOCF-SP retains the decomposition
\(f(\mathbf{x})=\epsilon(\mathbf{x})f^{P}(\mathbf{x})+\delta(\mathbf{x})\)
and the same correction rule, but places independent GP surrogates on
\(\epsilon(\mathbf{x})\), \(f^{P}(\mathbf{x})\), and \(\delta(\mathbf{x})\).
It thus retains composite modeling across these three components while treating the patch-level SSE as a single unknown function, without modeling its underlying patch summaries.
For all GP-based methods, we use a constant mean function and a Mat\'ern-$5/2$ covariance kernel, with hyperparameters estimated by maximum likelihood estimation (MLE). Each algorithm is initialized with the same 7-point Sobol design and run for a total of 50 evaluations. 
We repeat each experiment over 20 trials with different initial designs.
% We repeat each experiment over 20 independent trials with different initial designs. 
For SBOCF and BOCF-SP, we use the \(3\times3\) square partition shown in Fig.~\ref{fig:patch_pixel_sse}(b) and set the parameter \(c\) in Eq.~\eqref{eq:sbocf_decomposition_opt} to 1 for all problems considered.

% We measure optimization performance by the best (lowest) pixel-wise SSE observed by each iteration, averaged over the 20 trials. We also quantify the performance gain of SBOCF using the ratio of the median baseline metric to the corresponding median SBOCF metric, considering both the final SSE and absolute parameter-estimation error. A ratio greater than one indicates an improvement over the baseline. In addition, we report the average acquisition-function optimization time per iteration, excluding the PACBED simulation time, which is common to all methods.
We measure optimization performance by the best (lowest) pixel-wise SSE observed by each iteration, averaged over the 20 trials. We quantify the performance gain of SBOCF as the ratio of the median final SSE achieved by each baseline to that achieved by SBOCF. We also report the corresponding ratio for absolute parameter-estimation error. A ratio greater than one indicates an improvement over the baseline. In addition, we report the average acquisition-function optimization time per iteration, excluding the PACBED simulation time, which is common to all methods.

% All methods are implemented in Python using \emph{PyTorch}~\cite{paszke_pytorch_2019} and \emph{BoTorch}~\cite{balandat_botorch_2020}. The code is available at \url{https://github.com/dasol-yoon/bott}. Experiments are run on a single computing node with one NVIDIA GeForce RTX 3090 GPU and four CPU cores. See full optimization details in Appendix~\ref{app:optdetails}.
All methods are implemented in Python using \emph{PyTorch}~\cite{paszke_pytorch_2019} and \emph{BoTorch}~\cite{balandat_botorch_2020}. Our code is available at \url{https://github.com/dasol-yoon/bott}. Experiments run on one node with an NVIDIA GeForce RTX 3090 GPU and four CPU cores. Optimization details are provided in Appendix~\ref{app:optdetails}.

\subsection{Benchmarks}
\paragraph{Simulated SrTiO$_3$ PACBED.} 
We first evaluate the methods on synthetic SrTiO$_3$ PACBED benchmarks, where the target PACBED images are generated by the same simulator used during optimization. This setting provides a controlled test case in which the ground-truth specimen parameters are known and, in the noiseless case, the global optimum corresponds to the parameter setting that exactly reproduces the target image.

PACBED patterns are simulated on-the-fly using the abTEM package~\cite{madsen_abtem_2021} during the optimization process, with BO adaptively selecting the simulation variables. Each evaluation simulates a 4D-STEM scan over probe positions spanning a unit cell and averages the resulting diffraction patterns to form the PACBED pattern. We follow the forward simulation settings for SrTiO$_3$ considered in~\cite{xu_deep_2018}. Detailed simulation settings are provided in Appendix~\ref{app:simulation_settings}.
% Specifically, we use a beam energy of 200 keV, a convergence semi-angle of 19.1 mrad, a defocus of 0 nm, a collection angle of 31 mrad, a scan step size of \SI{0.3}{\angstrom}, and a potential extent of \SI{62.6}{\angstrom}. Each PACBED pattern has a resolution of $157\times157$ pixels.
% Specifically, we use a beam voltage of 200 keV, a convergence semi-angle of 19.1 mrad, a defocus of 0 nm, a collection angle of 31 mrad, a scan step size of \SI{0.3}{\angstrom}, and a potential extent of \SI{62.6}{\angstrom}. Under these settings, each PACBED pattern had a resolution of 157 $\mathrm{\times}$ 157 pixels.

The unknown parameters are the specimen thickness and the beam tilts along the \(x\)- and \(y\)-axes, searched over \([5,500]\)~\textup{\AA} and \([-10,10]\)~mrad, respectively. We consider three simulated benchmarks: \texttt{Sim380} and \texttt{Sim100}, thick and thin specimens with ground-truth thicknesses of 380 and 100~\textup{\AA} and tilts of \((1.5,-1.5)\)~mrad; and \texttt{Sim200}, at 200~\textup{\AA} and \((3,-5)\)~mrad. Results for \texttt{Sim100} and \texttt{Sim200} are deferred to Appendices~\ref{app:sim100} and~\ref{app:sim200}.
%which supplies the specimen parameters for the downstream ptychographic reconstruction in Section~\ref{sec:results}. The 4D-STEM dataset reconstructed there is simulated with the same forward model and microscope settings, so the \texttt{Sim200} estimates transfer to it directly. Results for \texttt{Sim100} and \texttt{Sim200} are deferred to Appendices~\ref{app:sim100} and~\ref{app:sim200}.

We additionally evaluate robustness to measurement noise by adding varying noise levels to the target images %in \texttt{Sim380} and \texttt{Sim100}
; see Appendix~\ref{app:simnoise}. We also compare the \(3\times3\) square partition with an alternative partition comprising a high-intensity circular center and four outer quadrants; see Appendix~\ref{app:newpattern}.

\paragraph{Experimental SrTiO$_3$ PACBED.}
We next evaluate the methods on an experimental SrTiO$_3$ PACBED pattern from~\cite{xu_deep_2018}, using the same simulation setup, optimization variables, and search space as for the simulated benchmarks. Unlike the simulated targets, the experimental pattern cannot be reproduced exactly by the simulator, and its true specimen parameters are unknown. We therefore evaluate the parameter estimates using the reference values reported in~\cite{xu_deep_2018}. Specifically, we use a thickness of $380$~\textup{\AA} and beam tilts of $(1.5,-1.5)$~mrad. We refer to this benchmark as \texttt{Exp380}.

\paragraph{Downstream MEP reconstruction.}
We use \texttt{Sim200} as a representative test case for downstream MEP reconstruction. The corresponding 4D-STEM dataset is generated using the same specimen parameters and microscope settings, and the SBOCF-estimated thickness and mistilt are then used in the reconstruction. 
\section{Results and discussion}\label{sec:results}
\paragraph{Simulated SrTiO$_3$ PACBED:}

\begin{figure}[h!]
    \centering
    \includegraphics[width=0.85\linewidth]{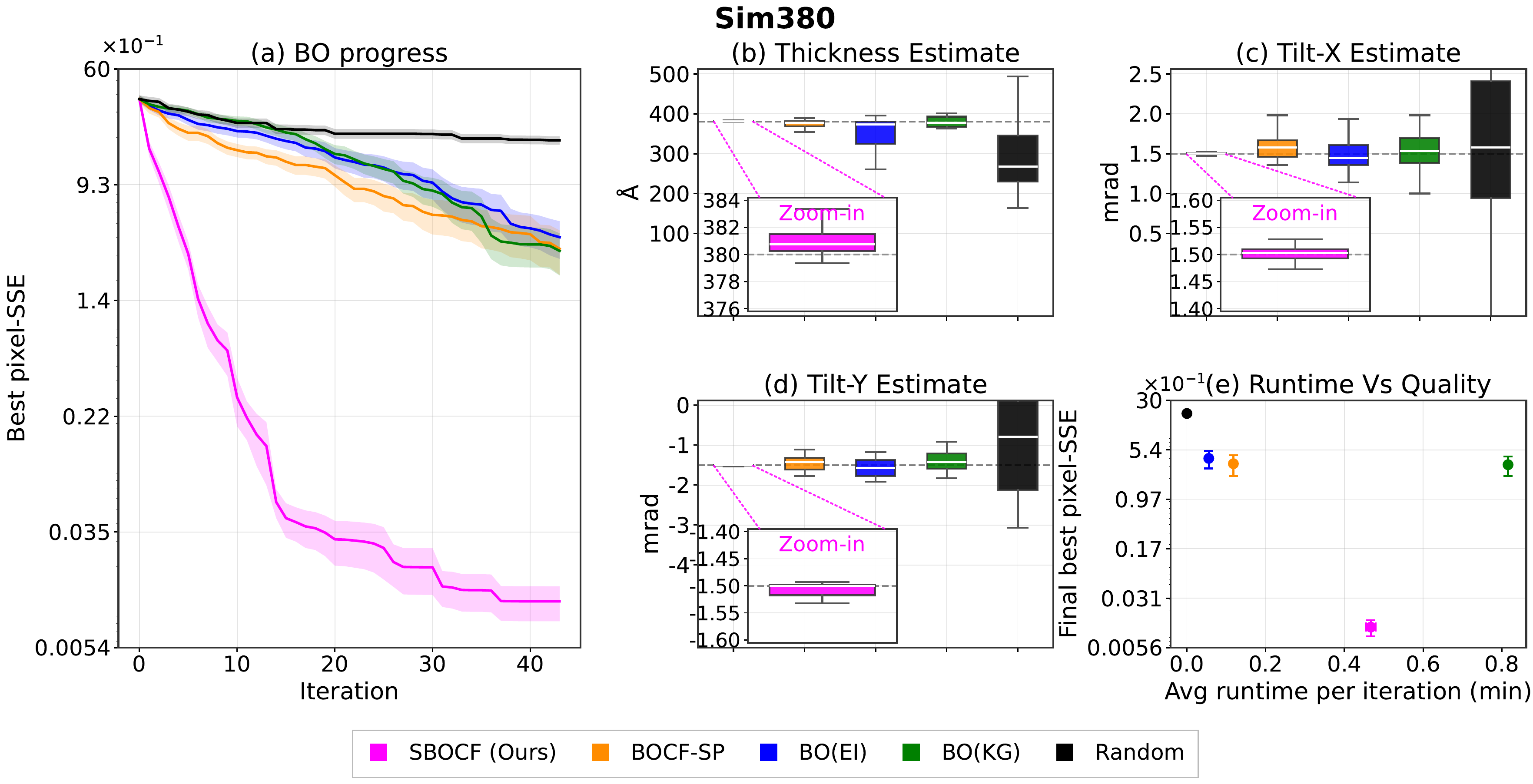}
    \caption{Results for \texttt{Sim380} with ground-truth thickness $380$~\AA{} and tilts $(1.5,-1.5)$~mrad. Panels show (a) best observed pixel-wise SSE, final estimates of (b) thickness, (c) tilt-X, and (d) tilt-Y, and (e) acquisition runtime versus final SSE. Solid lines in Panel~(a) show the mean across 20 trials, and the shaded regions indicate standard error. Dashed lines in Panels~(b)--(d) indicate the ground truth.}
    \label{fig:sim380}
\end{figure}

Fig.~\ref{fig:sim380} summarizes the results for \texttt{Sim380}; result figures for \texttt{Sim100} and \texttt{Sim200} are provided in Appendices~\ref{app:sim100} and~\ref{app:sim200}, respectively. In each figure, Panel~(a) shows the best observed pixel-wise SSE at each iteration; Panels~(b)--(d) show the final thickness, tilt-X, and tilt-Y estimates; and Panel~(e) compares the average acquisition runtime per iteration with the final SSE. Both standard BO baselines outperform random sampling, while SBOCF achieves the lowest final SSE and the most accurate and stable parameter estimates. BOCF-SP consistently underperforms SBOCF and is often comparable to or slightly worse than standard BO.

Relative to BO(EI), BO(KG), and BOCF-SP, respectively, SBOCF improves final SSE by factors of $(291,302,110)\times$ for \texttt{Sim380}, $(43,399,141)\times$ for \texttt{Sim100}, and $(49,214,123)\times$ for \texttt{Sim200}. For thickness, tilt-X, and tilt-Y errors, respectively, the improvement factors are listed below, with entries in each tuple ordered as BO(EI), BO(KG), and BOCF-SP: $(21,20,36)\times$, $(17,25,41)\times$, and $(9.5,20,33)\times$ for \texttt{Sim380}; $(1.5,12,7.7)\times$, $(5.5,29,23)\times$, and $(7.3,13,14)\times$ for \texttt{Sim100}; and $(1.7,5.3,7.1)\times$, $(5.5,9.0,12)\times$, and $(2.3,9.3,9.0)\times$ for \texttt{Sim200}. These results support the benefit of explicitly modeling patch summaries: BOCF-SP omits this structure while introducing additional model components through which estimation error can propagate.
% \pbedit{BOCF-SP is consistently worse than SBOCF and is often comparable to or slightly worse than standard BO.}

% \pbedit{For \texttt{Sim380}, SBOCF achieves final-SSE improvement factors of $291\times$, $302\times$, and $110\times$ over BO(EI), BO(KG), and BOCF-SP, respectively. For thickness, tilt-X, and tilt-Y, the corresponding improvement factors are $(21\times,20\times,36\times)$, $(17\times,25\times,41\times)$, and $(9.5\times,20\times,33\times)$. For \texttt{Sim100}, the final-SSE improvement factors are $43\times$, $399\times$, and $141\times$, with corresponding parameter-error improvements of $(1.5\times,12\times,7.7\times)$, $(5.5\times,29\times,23\times)$, and $(7.3\times,13\times,14\times)$. For \texttt{Sim200}, the final-SSE improvement factors are $49\times$, $214\times$, and $123\times$, with corresponding parameter-error improvements of $(1.7\times,5.3\times,7.1\times)$, $(5.5\times,9.0\times,12\times)$, and $(2.3\times,9.3\times,9.0\times)$. Across these settings, BOCF-SP lags SBOCF by a wide margin and often fails to improve upon standard BO. This is because BOCF-SP does not fully exploit the composite structure that SBOCF is designed to use: it relies on information similar to that available to standard BO, but through a more complicated model in which estimation error can propagate.}

SBOCF is moderately more expensive per iteration than BO(EI) because it models 11 GP outputs rather than one, but achieves substantially lower final SSE. It is also faster and more accurate than BO(KG), whose fantasy-model construction and high-dimensional one-shot optimization incur greater computational cost~\cite{balandat_botorch_2020}. Thus, SBOCF provides a favorable trade-off between runtime and parameter-estimation accuracy. BOCF-SP, by contrast, incurs greater modeling overhead without a commensurate gain in accuracy.

\begin{figure}[h!]
    \centering
    \includegraphics[width=0.85\linewidth]{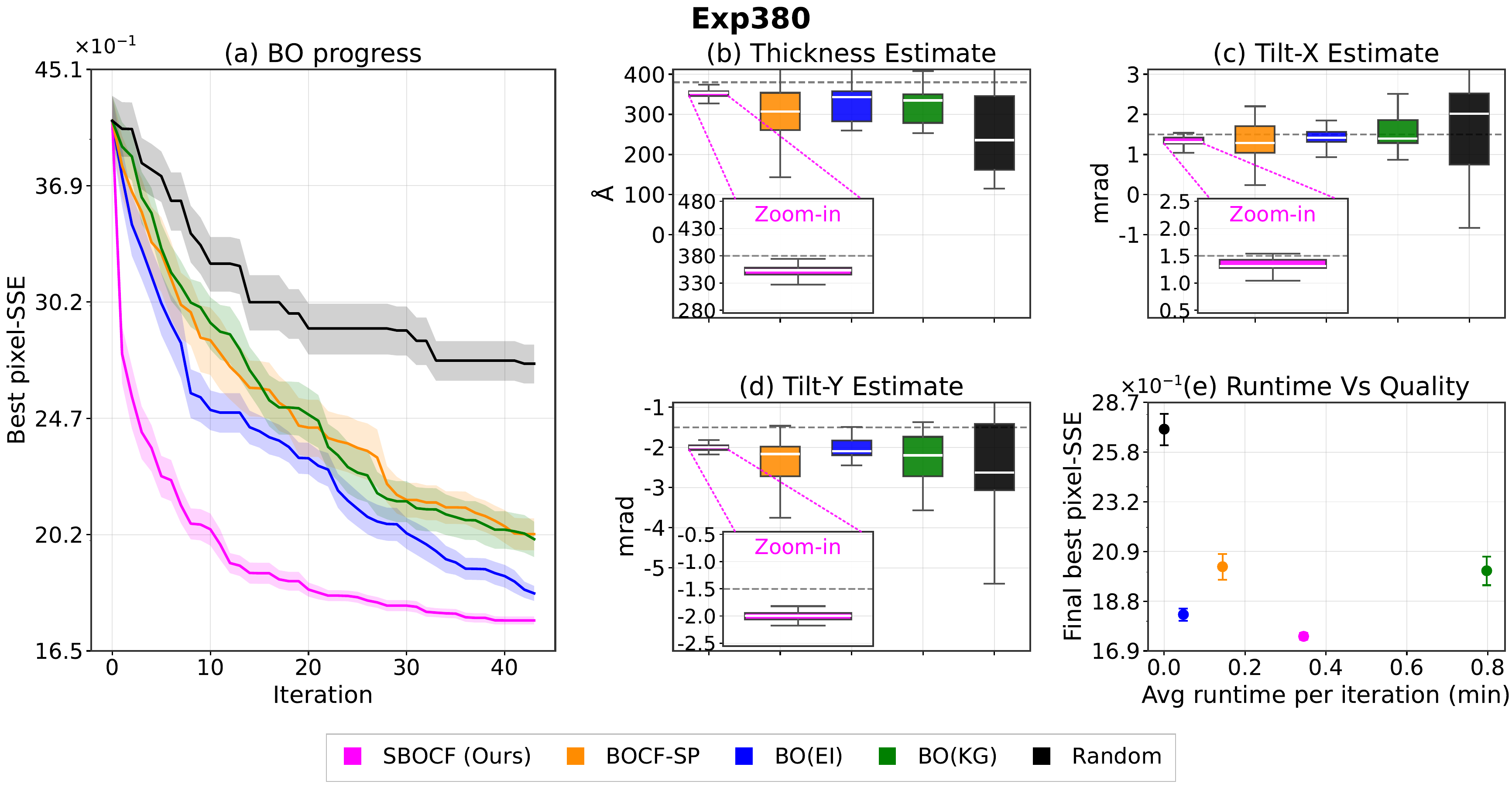}
    \caption{Results for the \texttt{Exp380} experimental SrTiO$_3$ PACBED benchmark using the target image from~\cite{xu_deep_2018}. Layout as in Fig.~\ref{fig:sim380}, except that dashed lines in Panels~(b)--(d) indicate the reference estimates reported in~\cite{xu_deep_2018}.}
    \label{fig:Exp380}
\end{figure}

\paragraph{Experimental SrTiO$_3$ PACBED:}
Fig.~\ref{fig:Exp380} shows \texttt{Exp380} results with the layout of Fig.~\ref{fig:sim380}. SBOCF achieves the lowest final SSE, improving over BO(EI), BO(KG), and BOCF-SP by factors of $1.03\times$, $1.11\times$, and $1.11\times$, respectively. Ratios of median absolute deviations from the reference estimates are $(1.5,0.81,1.2)\times$ for thickness, $(1.6,1.3,1.4)\times$ for tilt-X, and $(2.6,2.2,1.3)\times$ for tilt-Y, in the same baseline order. The smaller gains than in synthetic cases likely reflect measurement noise and simulation--experiment mismatch. Dashed lines mark earlier estimates obtained using a neural-network-based method~\cite{xu_deep_2018}; without known ground truth, agreement indicates consistency rather than accuracy. As in the synthetic studies, BOCF-SP underperforms SBOCF and does not improve upon standard BO.
% \paragraph{Experimental SrTiO$_3$ PACBED:}
% Fig.~\ref{fig:Exp380} summarizes the \texttt{Exp380} results using the same panel layout as Fig.~\ref{fig:sim380}. \pbedit{SBOCF achieves the lowest final SSE, with improvement factors of $1.03\times$, $1.11\times$, and $1.11\times$ over BO(EI), BO(KG), and BOCF-SP, respectively. For thickness, tilt-X, and tilt-Y, the corresponding factors are $(1.5\times,0.81\times,1.2\times)$, $(1.6\times,1.3\times,1.4\times)$, and $(2.6\times,2.2\times,1.3\times)$. These gains are smaller than in the synthetic cases because measurement noise and simulation--experiment mismatch make the experimental problem more challenging, but SBOCF still yields the strongest overall SSE and competitive parameter estimates. The dashed lines mark reference values from an earlier neural-network analysis~\cite{xu_deep_2018}; because Exp380 has no known ground truth, agreement with these values indicates consistency with that method rather than accuracy relative to a true specimen state. As in the synthetic studies, BOCF-SP lags SBOCF and does not improve upon standard BO.}

SBOCF achieves a lower final SSE than both BO(EI) and BO(KG), with greater acquisition-function overhead than BO(EI) but substantially less than BO(KG). It also yields experimental parameter estimates consistent with those from the earlier neural-network-based method, without requiring a large simulated training set or task-specific neural-network training.

\begin{figure}[h!]
    \centering
    \includegraphics[width=0.65\linewidth]{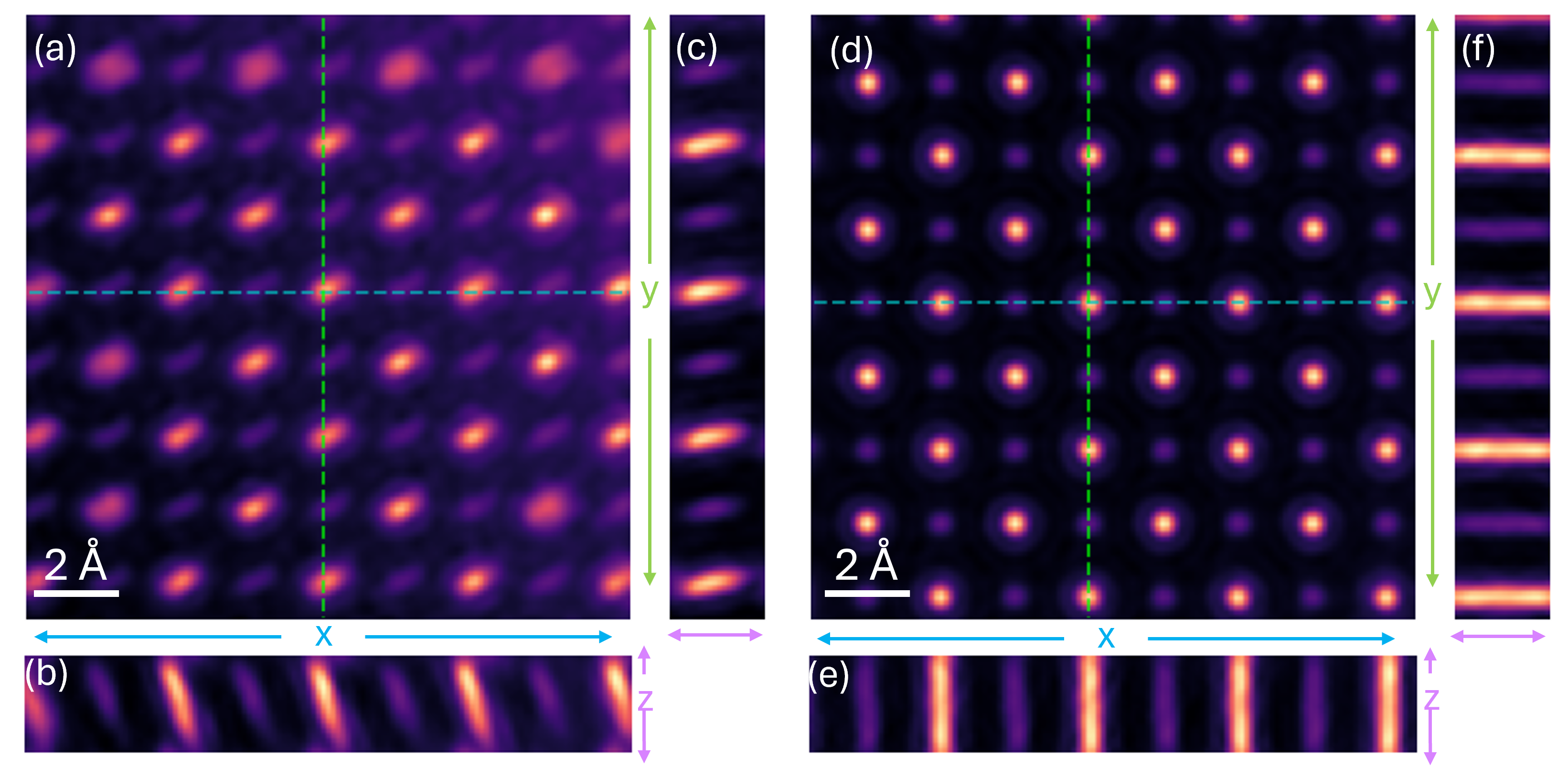}
    \caption{
    Effect of specimen mistilt on multislice electron ptychography and its correction using SBOCF estimates, for a simulated \ce{SrTiO3} specimen of thickness $200$~\AA{} with tilts of $3$~mrad (tilt-X) and $-5$~mrad (tilt-Y). (a)--(c) Reconstruction of the dataset without accounting for the tilt: (a) projected view, (b) $x$--$z$ and (c) $y$--$z$ depth sections. (d)--(f) The corresponding reconstruction using the SBOCF-estimated parameters. 
     % (\texttt{Sim200}; Appendix~\ref{app:sim200}). 
    %The tilt magnitudes are larger than the $(1.5,-1.5)$~mrad of the \texttt{Sim380} and \texttt{Sim100} benchmarks, so that the degradation is clearly visible. Scale bars, $2$~\AA{}.
    }
    \label{fig:mep}
\end{figure}

\paragraph{Downstream MEP reconstruction:} Fig.~\ref{fig:mep} illustrates the downstream consequence of accurate parameter estimates. For a simulated \ce{SrTiO3} specimen of thickness $200$~\AA{} with tilts of $3$ and $-5$~mrad, ignoring the tilt elongates the atomic columns---rows of atoms stacked along the beam direction, which appear as bright dots in a projected view---and tilts them with respect to the beam direction (a)--(c), whereas reconstructing with the SBOCF estimates recovers sharp, vertical columns (d)--(f). Those estimates match the ground truth to within $5$~\AA{} in thickness and $0.1$~mrad in tilt. 
% (\texttt{Sim200}; Appendix~\ref{app:sim200}).

\section{Conclusion}

% 20260824 CHL 8pm attempt

We introduced scalable Bayesian optimization of composite functions (SBOCF) for simulation-based estimation of specimen thickness and mistilt from PACBED images. By representing each simulated image with patch-level outputs and two correction terms, SBOCF exploits image-level information while preserving the pixel-wise SSE objective and reducing the number of modeled outputs by a factor of approximately $2{,}000$ relative to standard BOCF.

Under a limited simulation budget, SBOCF converged faster and achieved lower SSE than BO(EI), reducing the median final SSE by a factor of approximately $291$ on the noise-free synthetic \texttt{Sim380} dataset. On experimental data, SBOCF produced thickness and mistilt estimates consistent with previous neural-network-based estimates without requiring a large simulated training dataset.

Future work will extend SBOCF to additional PACBED-sensitive parameters, including probe aberrations and atomic-vibration
amplitudes. In addition, alternative
dimensionality reduction methods, such as principal component
analysis (PCA), could be investigated to assess whether low-rank representations offer
advantages over simple patch aggregation. More broadly, SBOCF
provides a practical approach to inverse problems involving expensive
simulations and high-dimensional structured outputs.

\begin{ack}
We would like to thank the anonymous reviewers for their constructive feedback and suggestions. This research was supported by the Center for Alkaline Based Energy Solutions (CABES), an Energy Frontier Research Center funded by the U.S. Department of Energy (DOE), Office of Science, Basic Energy Sciences (BES), under Award \#DE-SC0019445.  
PF and PB were partially supported by the AI Research Institutes program, supported by the NSF and the Intel Corporation under NSF Award DMR-2433348.
PB gratefully acknowledges financial support from the Development and Promotion of Science and Technology Talents Project (DPST) scholarship program of the Institute for the Promotion of Teaching Science and Technology (IPST), Ministry of Education, Thailand. CHL is supported by the Eric and Wendy Schmidt AI in Science Postdoctoral Fellowship, a program of Schmidt Sciences, LLC. 
\end{ack}

\bibliographystyle{unsrt}
\bibliography{references}

%%%%%%%%%%%%%%%%%%%%%%%%%%%%%%%%%%%%%%%%%%%%%%%%%%%%%%%%%%%%
\newpage
\appendix
\textbf{\Large Appendix}

\setcounter{figure}{0}
\setcounter{table}{0}
\setcounter{equation}{0}

\renewcommand{\thefigure}{\thesection\arabic{figure}}
\renewcommand{\thetable}{\thesection\arabic{table}}
\renewcommand{\theequation}{\thesection\arabic{equation}}

\counterwithin*{figure}{section}
\counterwithin*{table}{section}
\counterwithin*{equation}{section}

\section{Gaussian process posterior distribution}
\label{app:gpformulas}
Assume that 
$$f(\cdot)\sim\mathcal{GP}(\mu_0(\cdot),k_0(\cdot,\cdot)),$$
where $\mu_0(\cdot)$ is the prior mean function and $k_0(\cdot,\cdot)$ is the prior covariance function.

Given a dataset $\mathcal{D}_n=\{(\bx_i,f(\bx_i))\}_{i=1}^n$, conditioning on this data set, the posterior distribution of $f(\bx)$ follows~\cite{williams2006gaussian}:

$$f(\bx)|\mathcal{D}_n\sim\mathcal{N}(\mu_n(\bx),\sigma_n^2(\bx)),$$
where $\mu_n(\bx)$ and $\sigma_n^2(\bx)$ are posterior mean and posterior variance given by
$$\mu_n(\bx)=k_0(\bx,\mathbf{X})^TK(\mathbf{X},\mathbf{X})^{-1}(f(\mathbf{X})-\mu_0(\mathbf{X}))+\mu_0(\mathbf{x})$$
and
$$\sigma_n^2(\bx)=k_0(\bx,\bx)-k_0(\bx,\mathbf{X})^{T}K(\mathbf{X},\mathbf{X})^{-1}k_0(\bx,\mathbf{X}).$$

Here, $\mathbf{X}=\begin{bmatrix}
    \bx_1&\bx_2&\hdots&\bx_n
\end{bmatrix}$ is the matrix of $n$ evaluated inputs, $k_0(\bx,\mathbf{X})^T=\begin{bmatrix}
    k_0(\bx,\bx_1)&k_0(\bx,\bx_2)&\hdots&k_0(\bx,\bx_n)
\end{bmatrix}$ is a kernel vector between the queried point $\bx$ and evaluated points $\bx_i$ for $i=1,\hdots,n$, $K(\mathbf{X},\mathbf{X})$ is a $n\times n$ kernel matrix between $n$ evaluated points whose entries are given by $K(\mathbf{X},\mathbf{X})_{i,j}=k_0(\bx_i,\bx_j)$. $\mu_0(\mathbf{X})^T=\begin{bmatrix}
    \mu_0(\bx_1)&\mu_0(\bx_2)&\hdots&\mu_0(\bx_n)
\end{bmatrix}$ is a mean vector and $f(\mathbf{X})^T=\begin{bmatrix}
    f(\bx_1)&f(\bx_2)&\hdots&f(\bx_n)
\end{bmatrix}$ is an observation vector.
\section{Closed-form expected improvement formula}
\label{app:eiformula}
Given the Gaussian posterior distribution
$$f(\bx)|\mathcal{D}_n\sim\mathcal{N}(\mu_n(\bx),\sigma_n^2(\bx)),$$
EI admits the closed-form formula as follows:
\begin{equation}
\alpha^{\mathrm{EI}}_n(\bx)= (f^{\min}_n - \mu_n(\bx)) \Phi\!\left(\frac{f^{\min}_n - \mu_n(\bx)}{\sigma_n(\bx)}\right)
+ \sigma_n(\bx) \phi\!\left(\frac{f^{\min}_n - \mu_n(\bx)}{\sigma_n(\bx)}\right),\label{eqn:ei} 
\end{equation}
where $f_n^{\min}=\min_{i=1,\ldots,n}f(\bx_i)$ is the best objective value observed so far. $\Phi(\cdot)$ and $\phi(\cdot)$ denote the standard normal cumulative distribution function (CDF) and probability density function (PDF), respectively. Derivation can be found, for example, in \cite{watanabe2024derivation}.
\section{Closed-form solution for $\epsilon(\bx)$ and $\delta(\bx)$ decomposition}
\label{app:eps_del_decom}

For a fixed input $\bx$, consider the optimization problem
\begin{equation}
\begin{aligned}
\min_{\epsilon(\bx),\,\delta(\bx)} \quad & |\delta(\bx)| \\
\text{s.t.}\quad
& |\log_{10}\epsilon(\bx)| \leq c,\\
& f(\bx)=\epsilon(\bx)f^{\mathcal P}(\bx)+\delta(\bx),
\end{aligned}
\label{eqn:eps_delta_opt_app}
\end{equation}
where $c>0$ is a user-specified constant.

We derive the closed-form solution below.

From the constraint
\[
|\log_{10}\epsilon(\bx)| \leq c,
\]
we obtain
\[
10^{-c} \leq \epsilon(\bx) \leq 10^c.
\]
Using the equality constraint in~\eqref{eqn:eps_delta_opt_app}, we can eliminate $\delta(\bx)$:
\[
\delta(\bx) = f(\bx) - \epsilon(\bx) f^{\mathcal P}(\bx).
\]
Substituting this into the objective gives the equivalent one-variable problem
\begin{equation}
\min_{\epsilon(\bx) \in [10^{-c},\,10^c]}
\left|f(\bx)-\epsilon(\bx) f^{\mathcal P}(\bx)\right|.
\label{eqn:onevar_eps}
\end{equation}

We first consider the case $f^{\mathcal P}(\bx)>0$. In this case, minimizing
\[
|f(\bx)-\epsilon(\bx) f^{\mathcal P}(\bx)|
\]
is equivalent to choosing $\epsilon(\bx)$ as close as possible to the unconstrained minimizer
\[
\epsilon(\bx) = \frac{f(\bx)}{f^{\mathcal P}(\bx)}.
\]
Therefore, the constrained optimum is obtained by projecting $\frac{f(\bx)}{f^{\mathcal P}(\bx)}$ onto the interval $[10^{-c},\,10^c]$, i.e.,
\[
\epsilon^*(\bx)
=
\min\left\{10^c,\max\left\{10^{-c},\frac{f(\bx)}{f^{\mathcal P}(\bx)}\right\}\right\}.
\]
The corresponding optimal $\delta(\bx)$ is then recovered from the equality constraint:
\[
\delta^*(\bx)=f(\bx)-\epsilon^*(\bx)f^{\mathcal P}(\bx).
\]

Next, consider the case $f^{\mathcal P}(\bx)=0$. Then the equality constraint reduces to
\[
f(\bx)=\delta(\bx),
\]
so $\delta(\bx)$ is fixed and the objective no longer depends on $\epsilon(\bx)$. In this case, any $\epsilon(\bx)$ satisfying $10^{-c}\leq \epsilon(\bx)\leq 10^c$ is optimal. For convenience, we set
\[
\epsilon(\bx)=1, \qquad \delta(\bx)=f(\bx).
\]
\section{Knowledge gradient formula}
\label{app:KGformula}
Given the Gaussian predictive distribution 
$$f(\mathbf{x})|\mathcal{D}_n\sim\mathcal{N}(\mu_n(\mathbf{x}),\sigma^2_n(\mathbf{x})),$$
KG is constructed as follows:
\begin{equation}
\alpha^{\mathrm{KG}}_n(\bx)
= \min_{\bx' \in \mathcal{X}} \mu_n(\bx')-\mathbb{E}\left[
\min_{\bx' \in \mathcal{X}} \mu_{n+1}(\bx')
\;\middle|\; \bx_{n+1} = \bx
\right],
\label{eqn:kg_main}
\end{equation}
where the first term is the current best inferred value, and the second term represents the expected future best inferred SSE after evaluating at $\bx$. Thus, $\alpha_n^{\mathrm{KG}}(\bx)$ quantifies the expected reduction in the best inferred SSE resulting from the additional evaluation. Unlike EI, KG does not have an analytical formula and has to be evaluated using Monte Carlo approximation and optimized in a one-shot style~\cite{balandat_botorch_2020}.
\section{Optimization details}
\label{app:optdetails}

In this section, we provide optimization details for each method. All Bayesian optimization methods are implemented in \texttt{BoTorch}~\cite{balandat_botorch_2020}. 

For BO(EI), we use the built-in analytical implementation of log Expected Improvement (logEI) acquisition function, which has been shown to be substantially more numerically stable than the standard EI formulation~\cite{ament2023unexpected}. 
% The acquisition function is optimized using the default multi-start optimization procedure in \texttt{BoTorch} with 20 restarts and 100 raw samples.

For BO(KG), we follow the one-shot KG acquisition function implementation in \texttt{BoTorch}. In particular, Eq.~\eqref{eqn:kg_main} is approximated using 16 fantasy samples and optimized through a deterministic sample-average approximation enabled by the reparameterization trick and automatic differentiation. 
% The resulting optimization problem is solved using 20 restarts and 100 raw samples.

For SBOCF and BOCF-SP, we follow the implementation strategy from prior work on Bayesian optimization of composite functions~\cite{astudillo_bayesian_2019}. Since the acquisition function does not admit a closed-form expression, we use Monte Carlo approximation with 512 samples drawn from a Sobol quasi-Monte Carlo (QMC) sampler together with the reparameterization trick and automatic differentiation to enable gradient-based optimization. 
% The acquisition function is optimized with 50 restarts and 100 raw samples. Since the method involves compositions of Gaussian process models, the resulting acquisition landscape is substantially more complex, making the use of additional optimization effort reasonable.

All acquisition functions are optimized using multistart
L-BFGS-B through \texttt{BoTorch}, with 50 restarts initialized
from 100 raw samples.

\section{PACBED Simulation Settings}
\label{app:simulation_settings}

We generate PACBED patterns using multislice simulations in abTEM~\cite{madsen_abtem_2021}, following the SrTiO$_3$ simulation settings in~\cite{xu_deep_2018}. We use a beam energy of 200~keV, a convergence semi-angle of 19.1~mrad, a defocus of 0~nm, a collection angle of 31~mrad, a scan step size of 0.3~\AA, and a potential extent of 62.6~\AA. For each candidate thickness and mistilt, we simulate a 4D-STEM scan over probe positions spanning one unit cell and average the resulting diffraction patterns to obtain a $157\times157$-pixel PACBED image.
\section{Figure for \texttt{Sim100}}\label{app:sim100}

\begin{figure}[h!]
    \centering
    \includegraphics[width=0.9\linewidth]{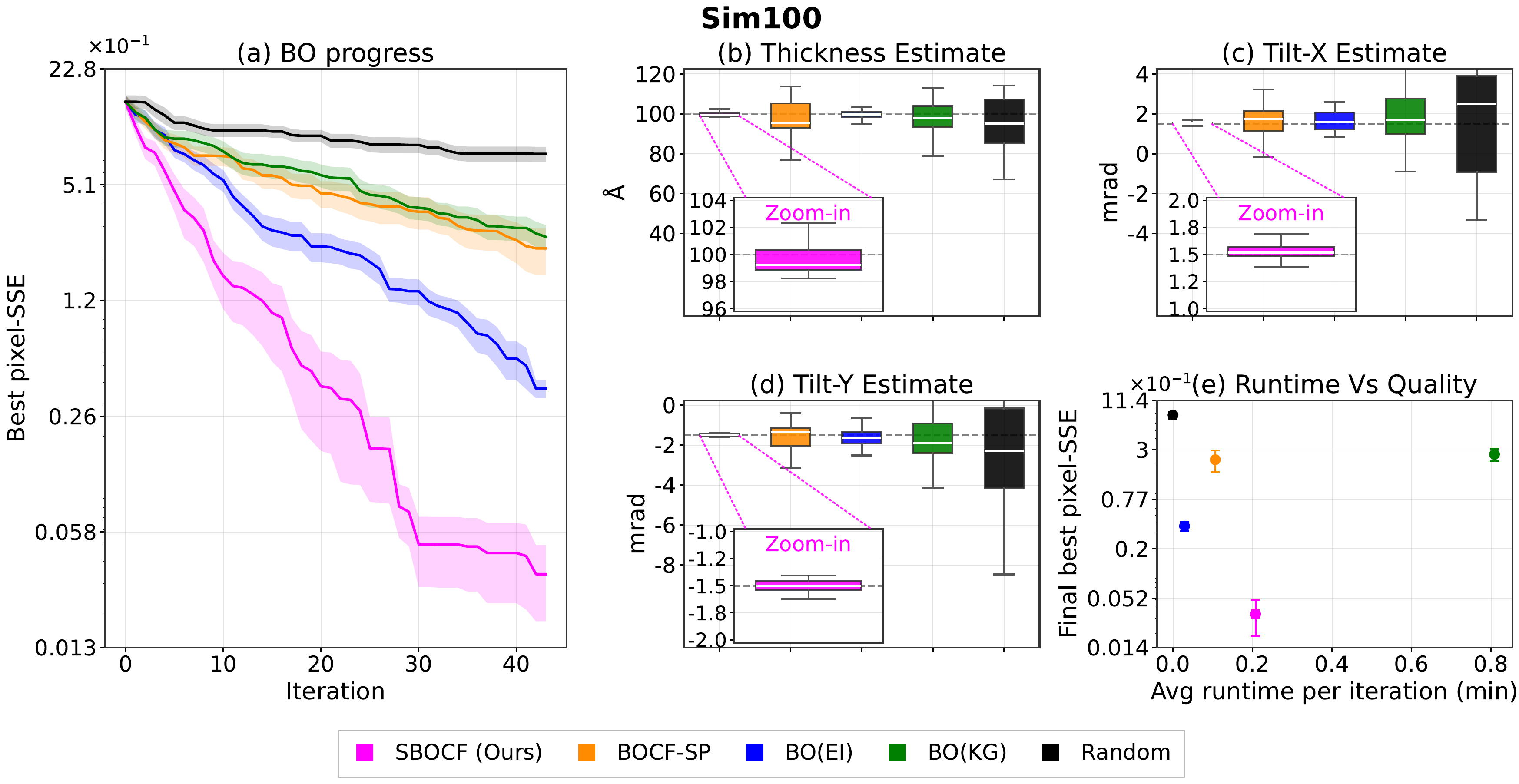}
    \caption{Results for \texttt{Sim100} with ground-truth thickness $100$~\AA{} and tilts $(1.5,-1.5)$~mrad. Panels show (a) best observed pixel-wise SSE, final estimates of (b) thickness, (c) tilt-X, and (d) tilt-Y, and (e) acquisition runtime versus final SSE. Solid lines in Panel~(a) show the mean across 20 trials, and the shaded regions indicate standard error. Dashed lines in Panels~(b)--(d) indicate the ground truth.}
    \label{fig:sim100}
\end{figure}

Fig.~\ref{fig:sim100} shows the \texttt{Sim100} results referred to in Section~\ref{sec:results}; the corresponding improvement factors are reported there.

\clearpage
\section{Figure for \texttt{Sim200}}\label{app:sim200}

% In this section we report results for an additional simulated benchmark, \texttt{Sim200}, with ground-truth thickness $200$~\AA{} and tilts of $3$ and $-5$~mrad. These parameters match the specimen used for the ptychographic reconstruction in Fig.~\ref{fig:mep}, so the estimates reported here are the ones supplied to that reconstruction. The tilt magnitudes are larger than the $(1.5,-1.5)$~mrad of \texttt{Sim380} and \texttt{Sim100}, which were chosen for comparability with the neural-network results of~\cite{xu_deep_2018}.

% We follow the same protocol as the main experiments: every method is initialized with the same 7-point Sobol design and run for 50 evaluations, repeated over 20 independent trials, using the $3\times3$ square partition and $c=1$ for SBOCF.

\begin{figure}[h!]
    \centering
    \includegraphics[width=0.9\linewidth]{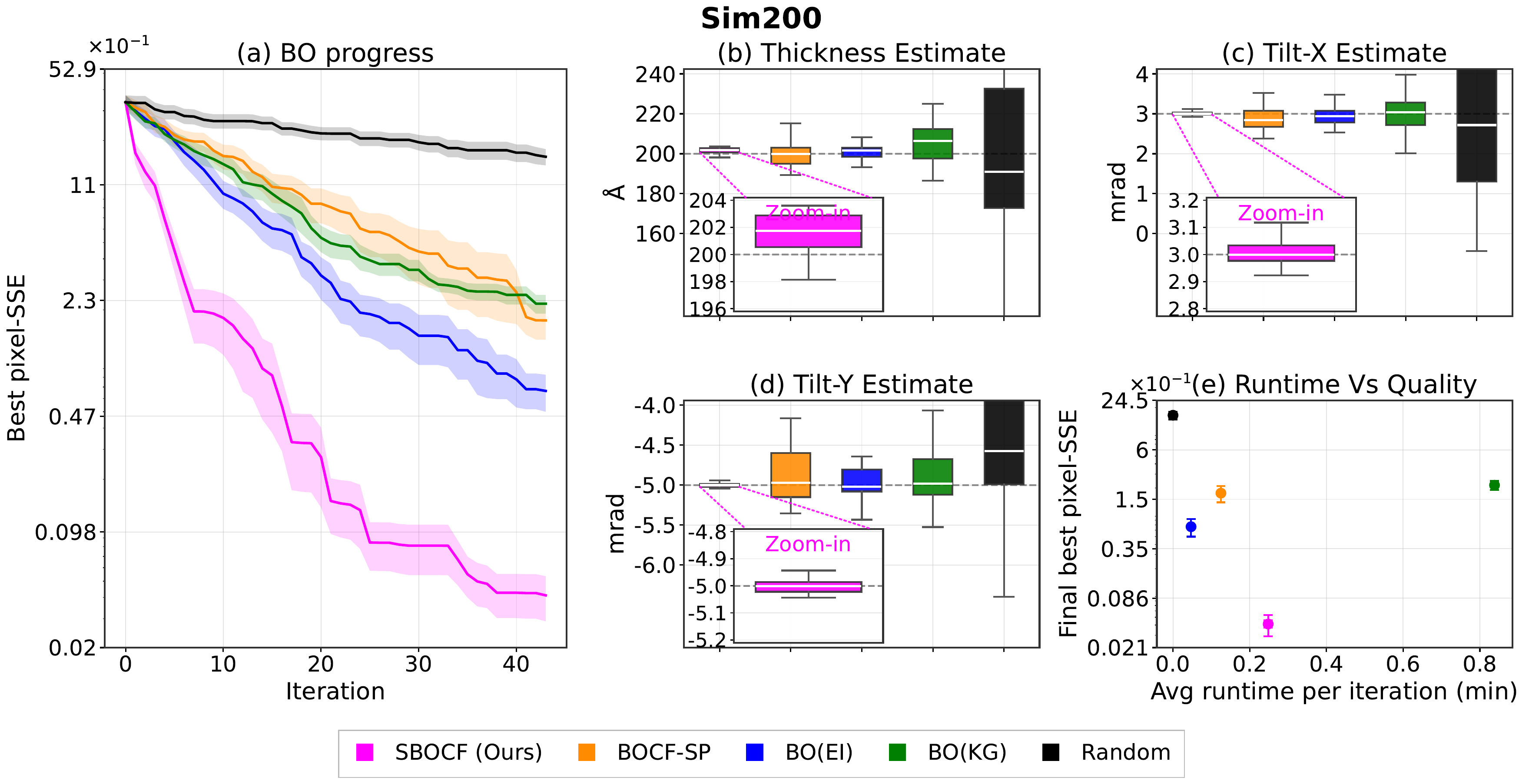}
    \caption{
Results for \texttt{Sim200} with ground-truth thickness $200$~\AA{} and tilts $(3,-5)$~mrad. Panels show (a) best observed pixel-wise SSE, final estimates of (b) thickness, (c) tilt-X, and (d) tilt-Y, and (e) acquisition runtime versus final SSE. Solid lines in Panel~(a) show the mean across 20 trials, and the shaded regions indicate standard error. Dashed lines in Panels~(b)--(d) indicate the ground truth.
}
    \label{fig:sim200}
\end{figure}

% Fig.~\ref{fig:sim200} summarizes the results using the same panel layout as Figs.~\ref{fig:sim380} and~\ref{fig:sim100}. SBOCF achieves the lowest final SSE, 
%with improvement factors of \textcolor{red}{XX}$\times$ and \textcolor{red}{YY}$\times$ over BO(EI) and BO(KG), respectively, 
% Fig.~\ref{fig:sim200} shows the \texttt{Sim200} results referred to in Section~\ref{sec:results}; the corresponding improvement factors are reported there. As we use this problem as a representative for downstream ptychrography application, we report that SBOCF recovers the ground-truth parameters to within $5$~\AA{} in thickness and $0.1$~mrad in tilt.
Fig.~\ref{fig:sim200} shows the \texttt{Sim200} results referred to in Section~\ref{sec:results}; the corresponding improvement factors are reported there. Since we use this problem as a representative case for the downstream ptychography application, we additionally note that SBOCF recovers the ground-truth parameters to within $5$~\AA{} in thickness and $0.1$~mrad in tilt.

\section{Ablation study: Simulated test cases with ground-truth noise}\label{app:simnoise}

In this section, we consider the simulated benchmark problems introduced in the main paper, namely \texttt{Sim380}, \texttt{Sim100} and \texttt{Sim200}. To better reflect realistic experimental conditions where measurements are corrupted by noise and the simulator cannot perfectly reproduce the observed image, we conduct an additional ablation study in which different levels of noise are added to the ground-truth PACBED images.

Specifically, we add Poisson noise to the ground-truth images after normalizing the image intensities and scaling them to simulated photon counts with a specified peak count parameter. Poisson-distributed counts are then sampled independently for each pixel before the noisy image is rescaled back to the original intensity range. For each problem, we consider three noise settings with peak photon count values of 150, 100, and 50, resulting in the benchmark problems \texttt{Sim380-P150}, \texttt{Sim380-P100}, \texttt{Sim380-P50}, \texttt{Sim100-P150}, \texttt{Sim100-P100},  \texttt{Sim100-P50},
\texttt{Sim200-P150}, \texttt{Sim200-P100}, and \texttt{Sim200-P50}. Smaller peak photon count values correspond to noisier observations due to larger relative Poisson fluctuations.

\begin{figure}[h!]
    \centering
    \includegraphics[width=\linewidth]{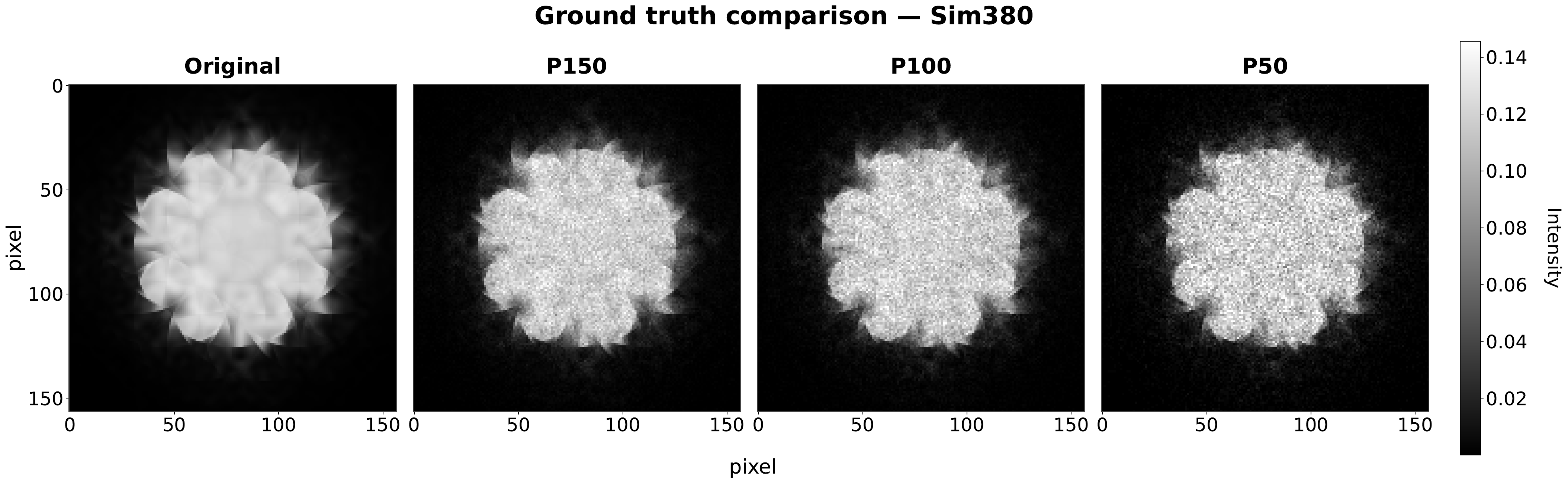}
    \caption{Comparison between the original ground truth PACBED pattern and noisy observations for the \texttt{Sim380} benchmark problem with a sample thickness of 380~\AA{}. From left to right, the figure shows the original noise-free image and noisy versions generated using peak photon counts of P150, P100, and P50. Lower peak photon counts correspond to higher levels of Poisson noise.}
    \label{fig:380compare}
\end{figure}

\begin{figure}[h!]
    \centering
    \includegraphics[width=\linewidth]{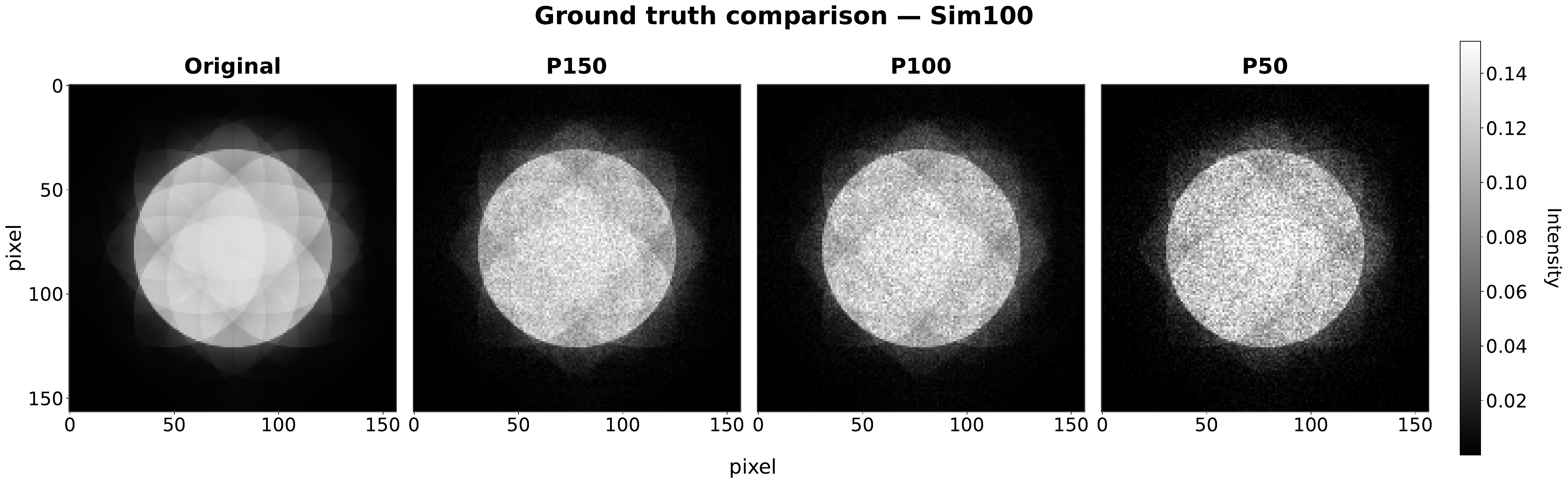}
\caption{Comparison between the original ground truth PACBED pattern and noisy observations for the \texttt{Sim100} benchmark problem with a sample thickness of 100~\AA{}. From left to right, the figure shows the original noise-free image and noisy versions generated using peak photon counts of P150, P100, and P50. Lower peak photon counts correspond to higher levels of Poisson noise.}
    \label{fig:100compare}
\end{figure}

\begin{figure}[h!]
    \centering
    \includegraphics[width=\linewidth]{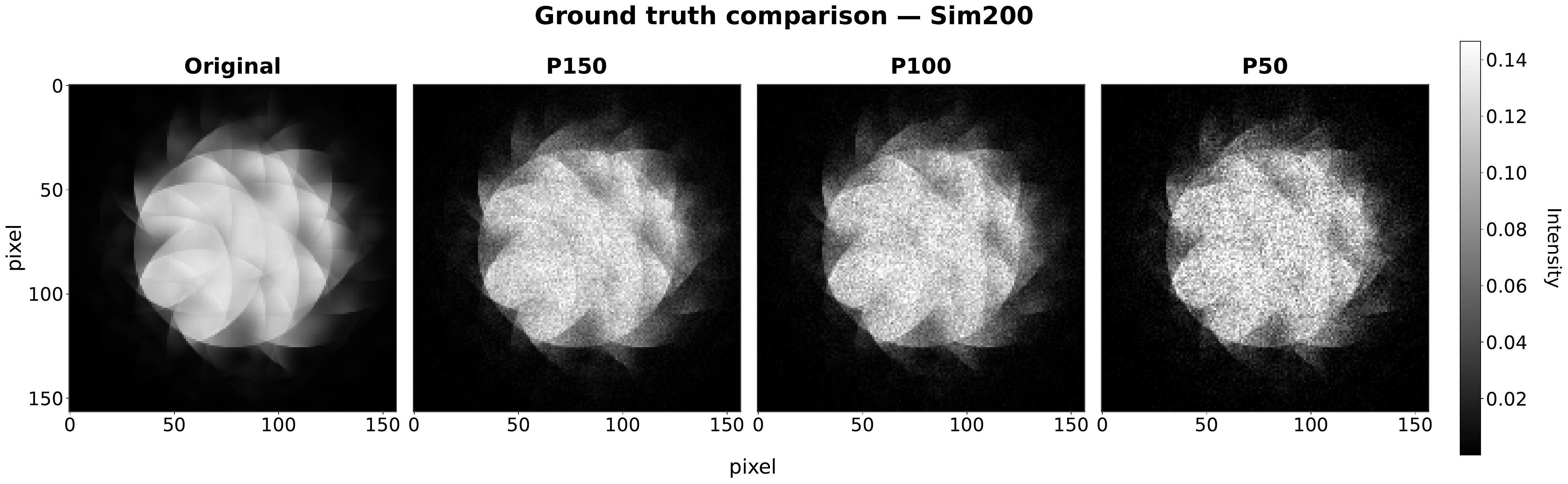}
\caption{Comparison between the original ground truth PACBED pattern and noisy observations for the \texttt{Sim200} benchmark problem with a sample thickness of 200~\AA{}. From left to right, the figure shows the original noise-free image and noisy versions generated using peak photon counts of P150, P100, and P50. Lower peak photon counts correspond to higher levels of Poisson noise.}
    \label{fig:200compare}
\end{figure}

The original ground-truth PACBED patterns and their noisy counterparts under different Poisson noise levels are shown in Fig.~\ref{fig:380compare} for \texttt{Sim380} (380~\AA{}), Fig.~\ref{fig:100compare} for \texttt{Sim100} (100~\AA{}), and Fig.~\ref{fig:200compare} for \texttt{Sim200} (200~\AA{}). Each figure presents the original noise-free ground truth image together with three noisy versions generated using peak photon count levels of P150, P100, and P50.

\begin{figure}[h!]
    \centering
    \includegraphics[width=0.9\linewidth]{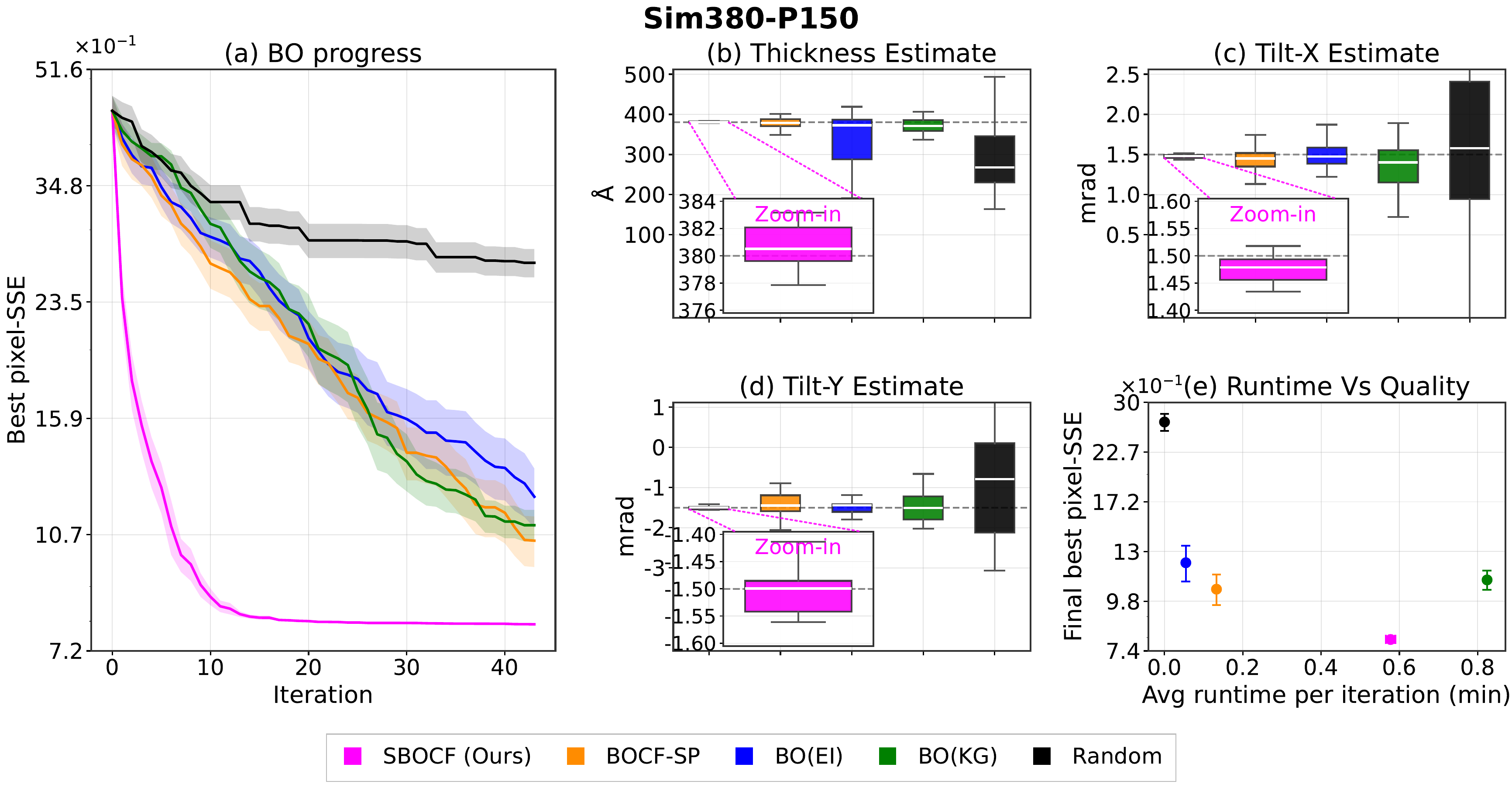}
\caption{
Results for \texttt{Sim380-P150} with ground-truth thickness $380$~\AA{}, tilts $(1.5,-1.5)$~mrad, and Poisson noise with a peak photon count of $150$. Panels show (a) best observed pixel-wise SSE, final estimates of (b) thickness, (c) tilt-X, and (d) tilt-Y, and (e) acquisition runtime versus final SSE. Solid lines in Panel~(a) show the mean across 20 trials, and the shaded regions indicate standard error. Dashed lines in Panels~(b)--(d) indicate the ground truth.
}
    \label{fig:sim380-p150}
\end{figure}

\begin{figure}[h!]
    \centering
    \includegraphics[width=0.9\linewidth]{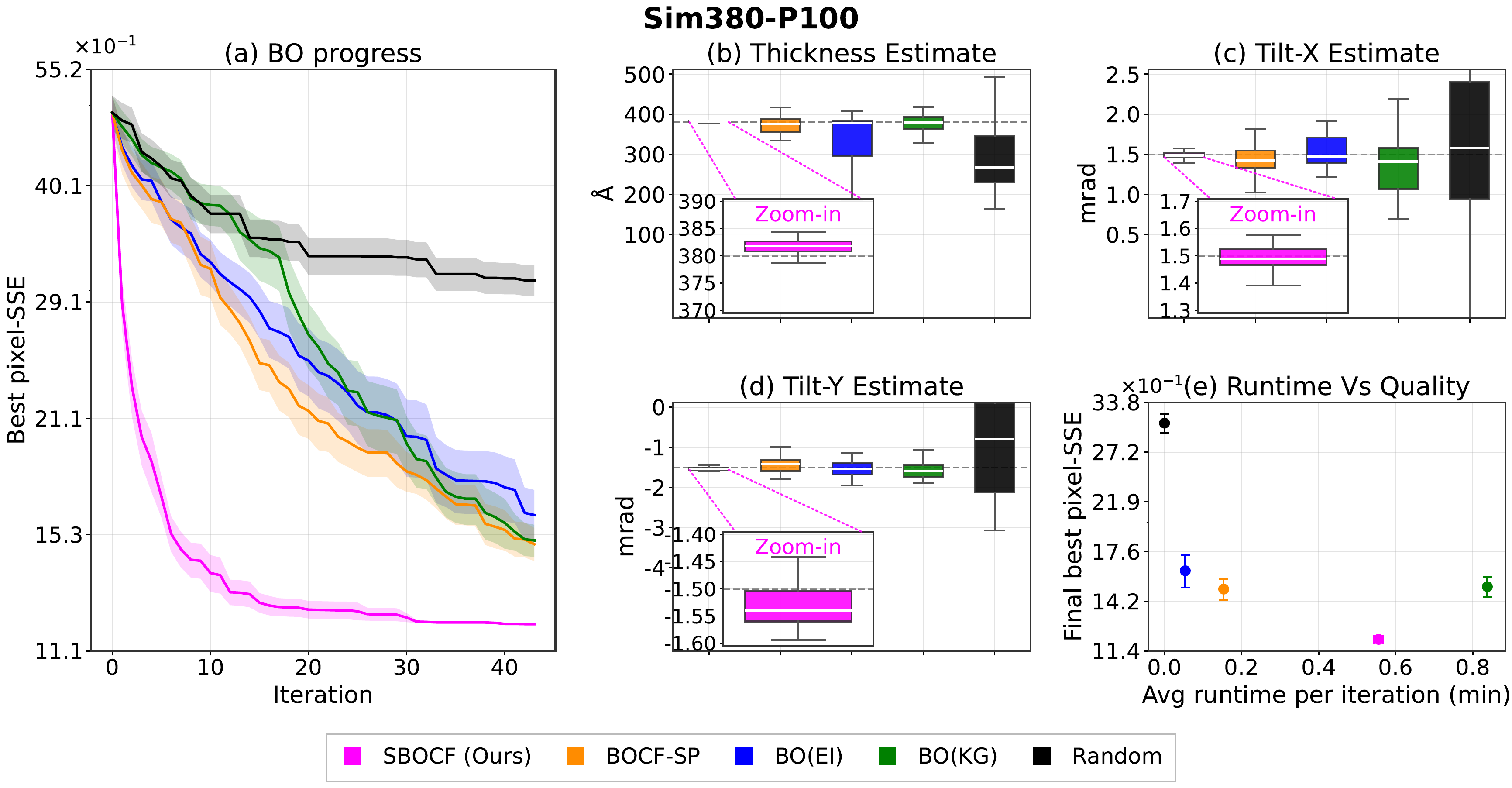}
\caption{
Results for \texttt{Sim380-P100} with ground-truth thickness $380$~\AA{}, tilts $(1.5,-1.5)$~mrad, and Poisson noise with a peak photon count of $100$. Panels show (a) best observed pixel-wise SSE, final estimates of (b) thickness, (c) tilt-X, and (d) tilt-Y, and (e) acquisition runtime versus final SSE. Solid lines in Panel~(a) show the mean across 20 trials, and the shaded regions indicate standard error. Dashed lines in Panels~(b)--(d) indicate the ground truth.
}
    \label{fig:sim380-p100}
\end{figure}

\begin{figure}[h!]
    \centering
    \includegraphics[width=0.9\linewidth]{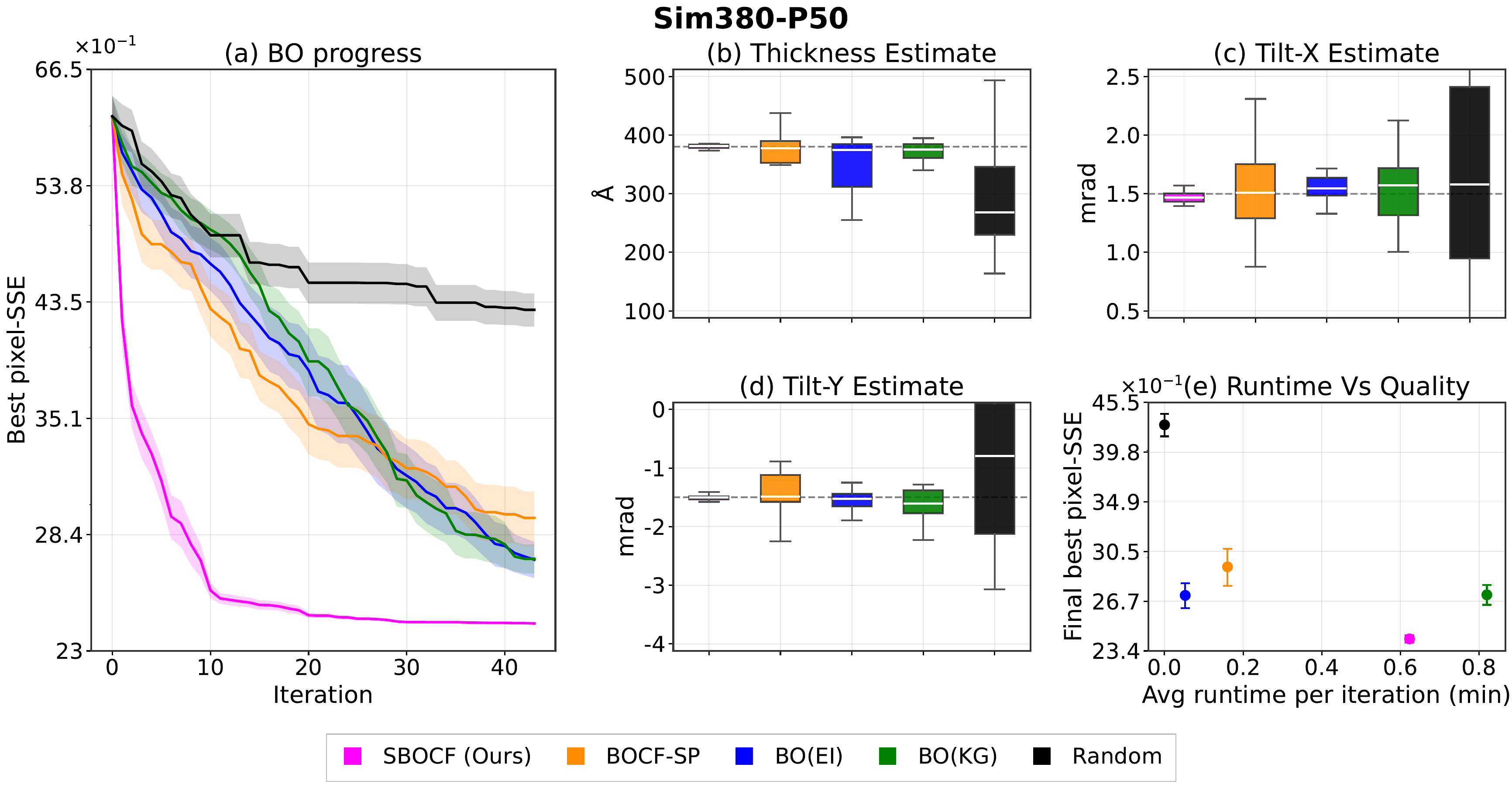}
\caption{
Results for \texttt{Sim380-P50} with ground-truth thickness $380$~\AA{}, tilts $(1.5,-1.5)$~mrad, and Poisson noise with a peak photon count of $50$. Panels show (a) best observed pixel-wise SSE, final estimates of (b) thickness, (c) tilt-X, and (d) tilt-Y, and (e) acquisition runtime versus final SSE. Solid lines in Panel~(a) show the mean across 20 trials, and the shaded regions indicate standard error. Dashed lines in Panels~(b)--(d) indicate the ground truth.
}
    \label{fig:sim380-p50}
\end{figure}

\begin{figure}[h!]
    \centering
    \includegraphics[width=0.9\linewidth]{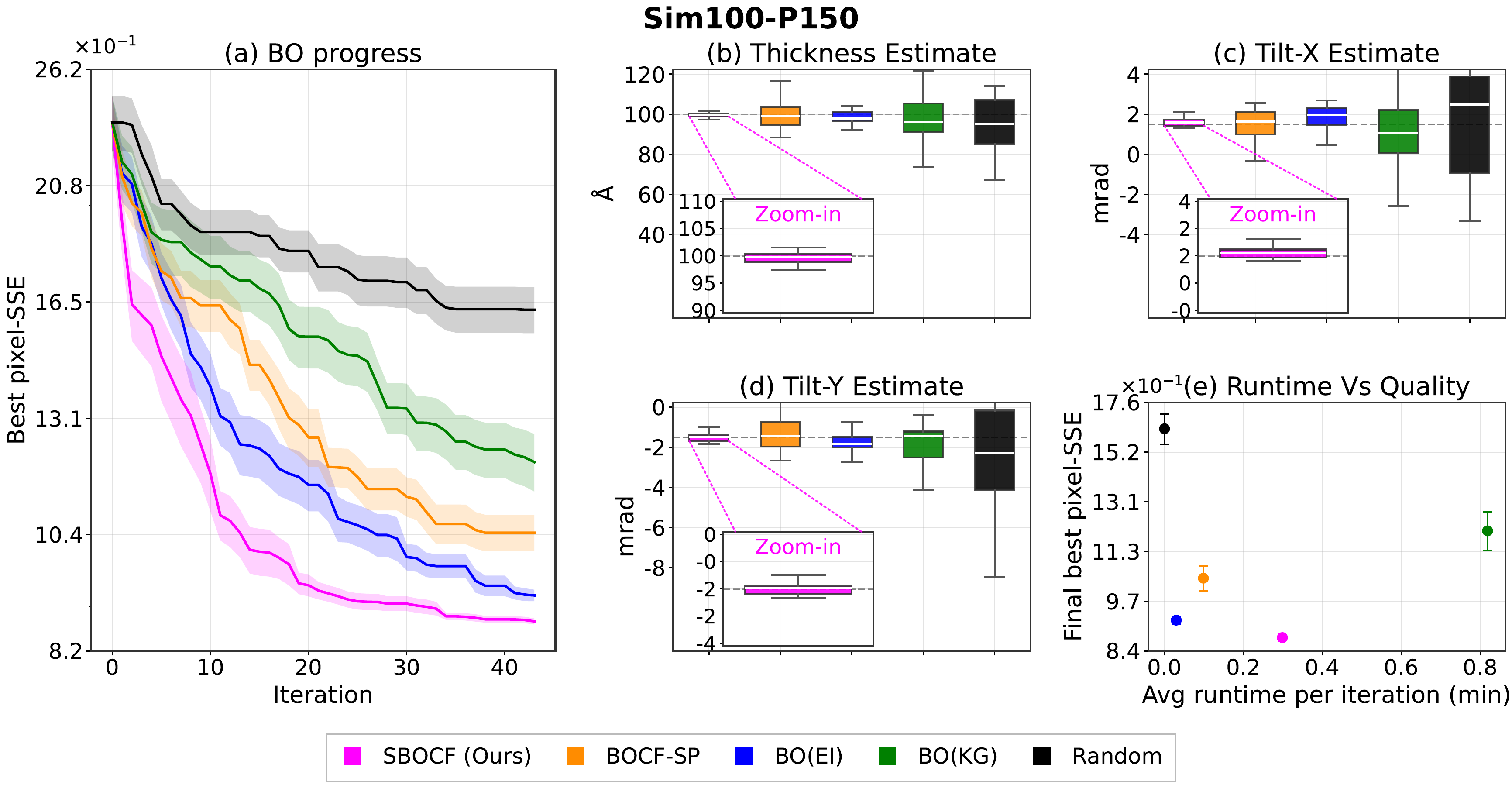}
\caption{
Results for \texttt{Sim100-P150} with ground-truth thickness $100$~\AA{}, tilts $(1.5,-1.5)$~mrad, and Poisson noise with a peak photon count of $150$. Panels show (a) best observed pixel-wise SSE, final estimates of (b) thickness, (c) tilt-X, and (d) tilt-Y, and (e) acquisition runtime versus final SSE. Solid lines in Panel~(a) show the mean across 20 trials, and the shaded regions indicate standard error. Dashed lines in Panels~(b)--(d) indicate the ground truth.
}
    \label{fig:sim100-p150}
\end{figure}

\begin{figure}[h!]
    \centering
    \includegraphics[width=0.9\linewidth]{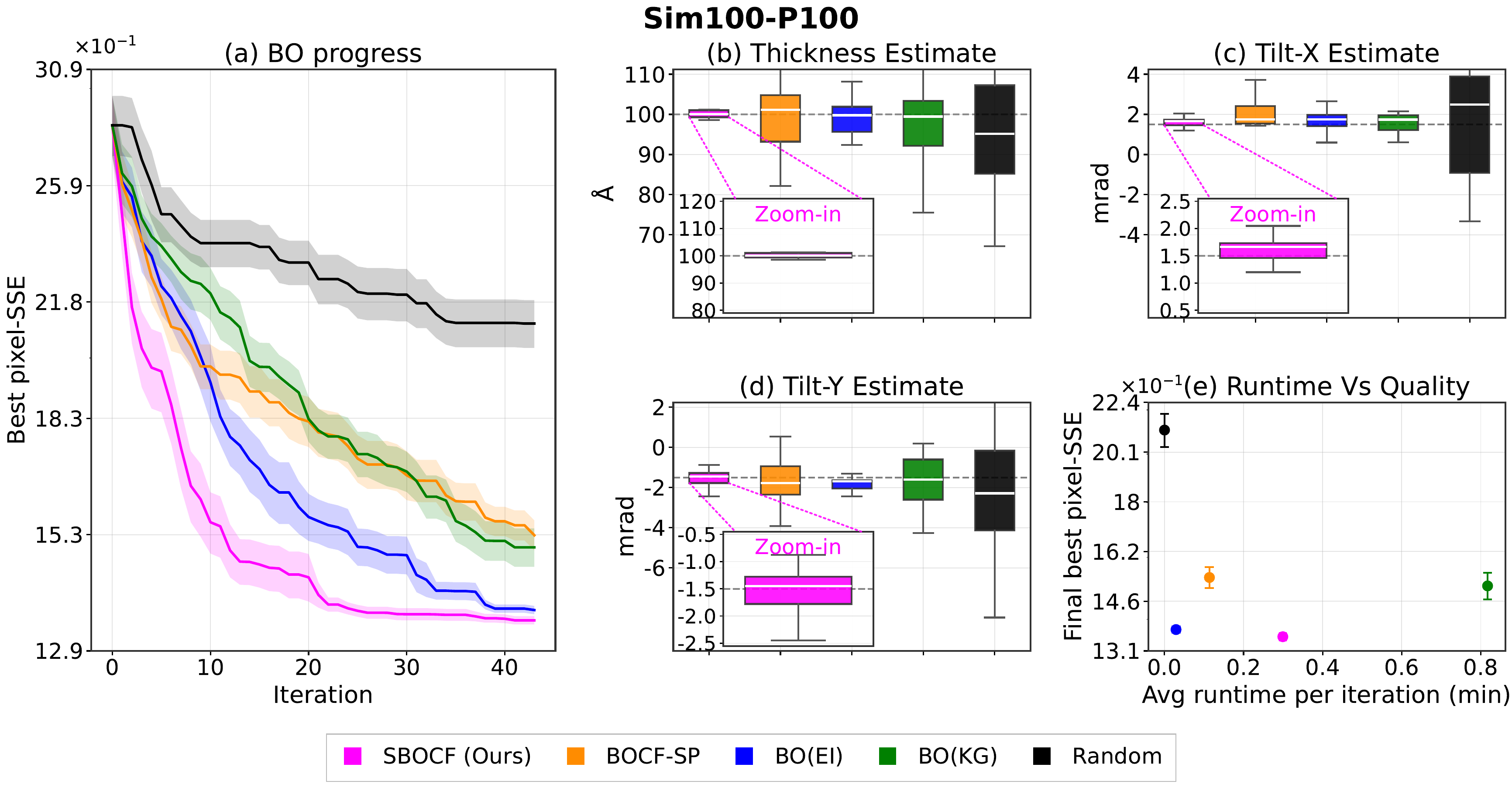}
\caption{
Results for \texttt{Sim100-P100} with ground-truth thickness $100$~\AA{}, tilts $(1.5,-1.5)$~mrad, and Poisson noise with a peak photon count of $100$. Panels show (a) best observed pixel-wise SSE, final estimates of (b) thickness, (c) tilt-X, and (d) tilt-Y, and (e) acquisition runtime versus final SSE. Solid lines in Panel~(a) show the mean across 20 trials, and the shaded regions indicate standard error. Dashed lines in Panels~(b)--(d) indicate the ground truth.
}
    \label{fig:sim100-p100}
\end{figure}

\begin{figure}[h!]
    \centering
    \includegraphics[width=0.9\linewidth]{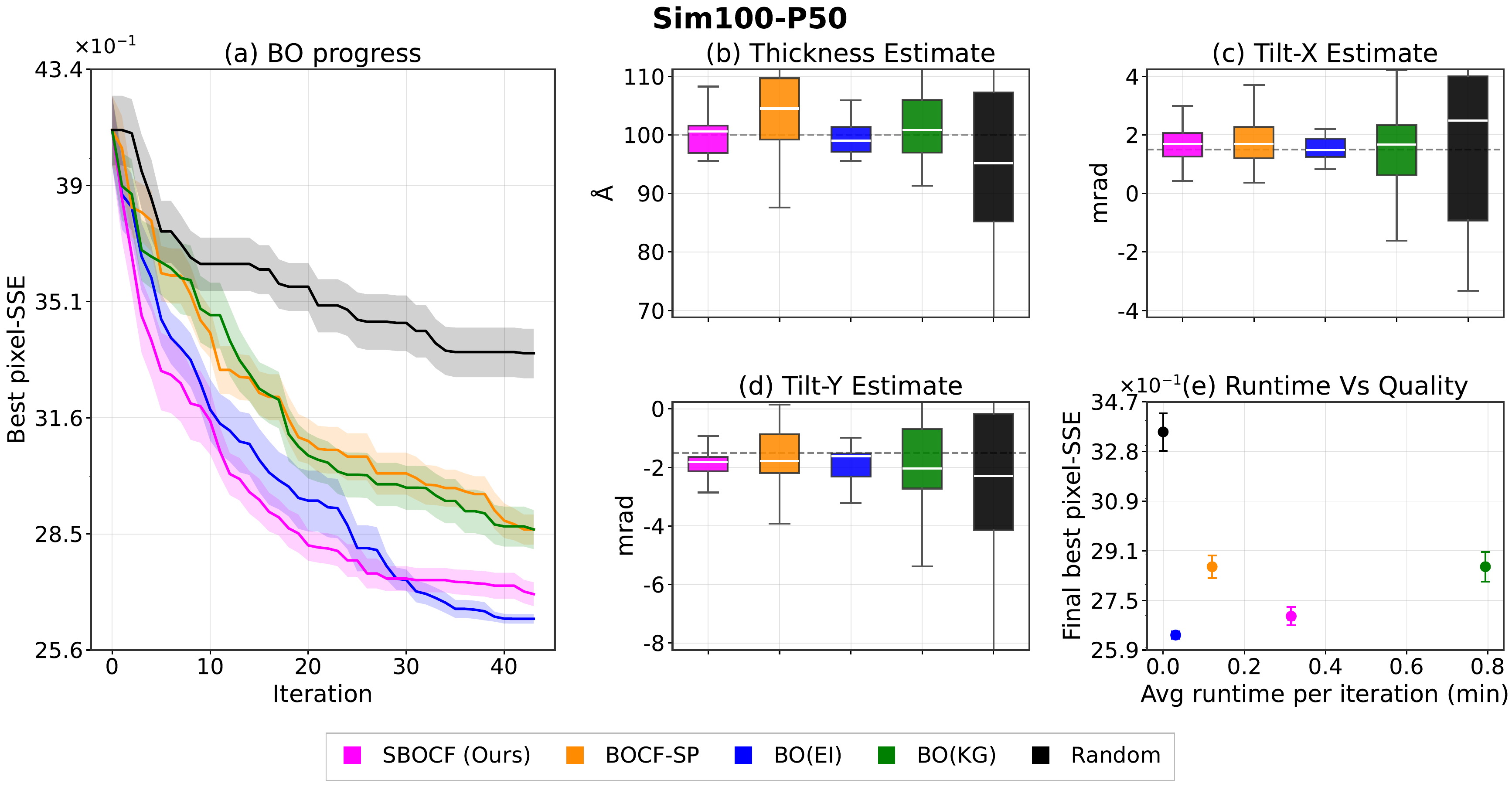}
\caption{
Results for \texttt{Sim100-P50} with ground-truth thickness $100$~\AA{}, tilts $(1.5,-1.5)$~mrad, and Poisson noise with a peak photon count of $50$. Panels show (a) best observed pixel-wise SSE, final estimates of (b) thickness, (c) tilt-X, and (d) tilt-Y, and (e) acquisition runtime versus final SSE. Solid lines in Panel~(a) show the mean across 20 trials, and the shaded regions indicate standard error. Dashed lines in Panels~(b)--(d) indicate the ground truth.
}
    \label{fig:sim100-p50}
\end{figure}

\begin{figure}[h!]
    \centering
    \includegraphics[width=0.9\linewidth]{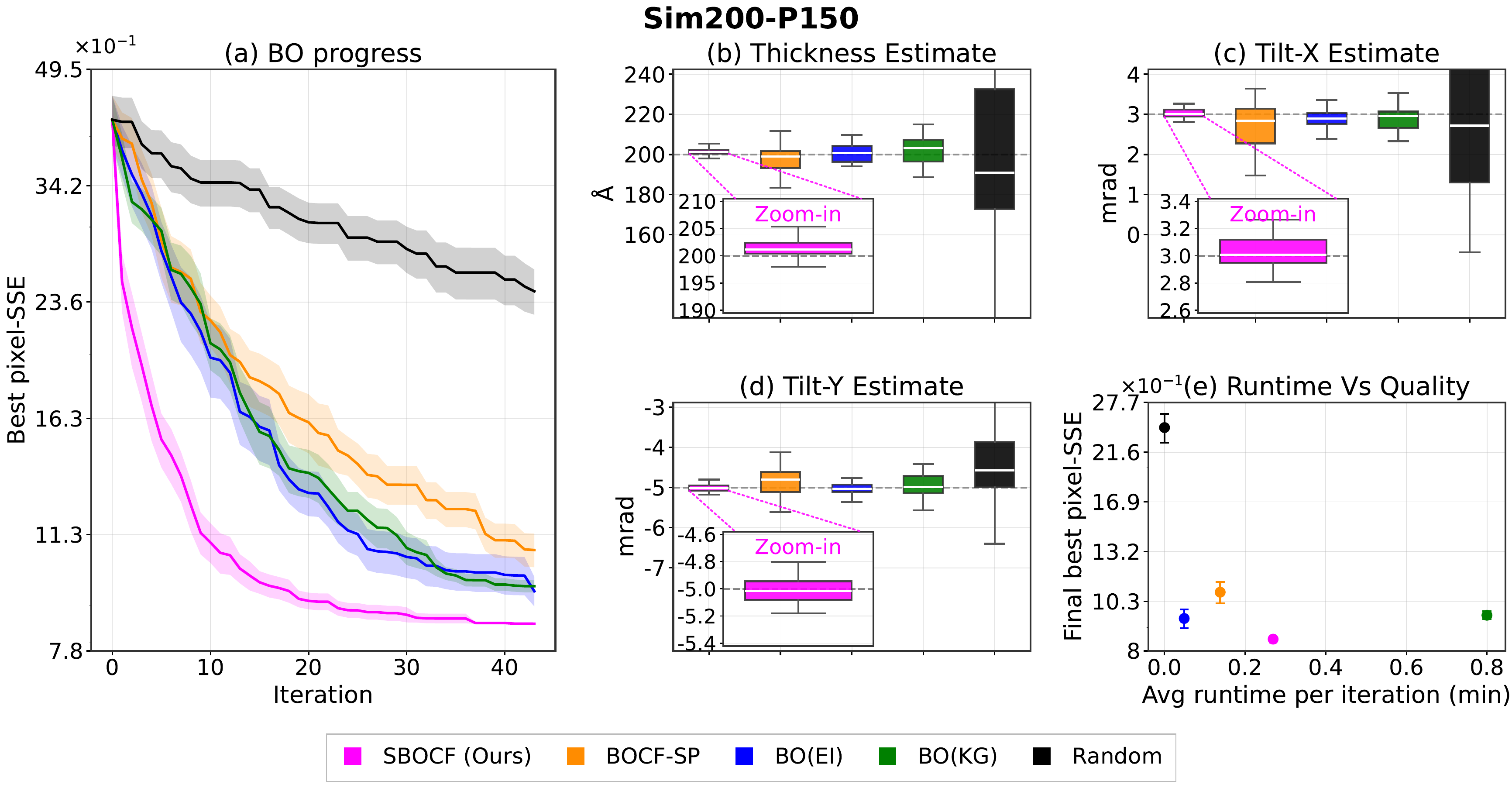}
\caption{
Results for \texttt{Sim200-P150} with ground-truth thickness $200$~\AA{}, tilts $(3,-5)$~mrad, and Poisson noise with a peak photon count of $150$. Panels show (a) best observed pixel-wise SSE, final estimates of (b) thickness, (c) tilt-X, and (d) tilt-Y, and (e) acquisition runtime versus final SSE. Solid lines in Panel~(a) show the mean across 20 trials, and the shaded regions indicate standard error. Dashed lines in Panels~(b)--(d) indicate the ground truth.
}
    \label{fig:sim200-p150}
\end{figure}

\begin{figure}[h!]
    \centering
    \includegraphics[width=0.9\linewidth]{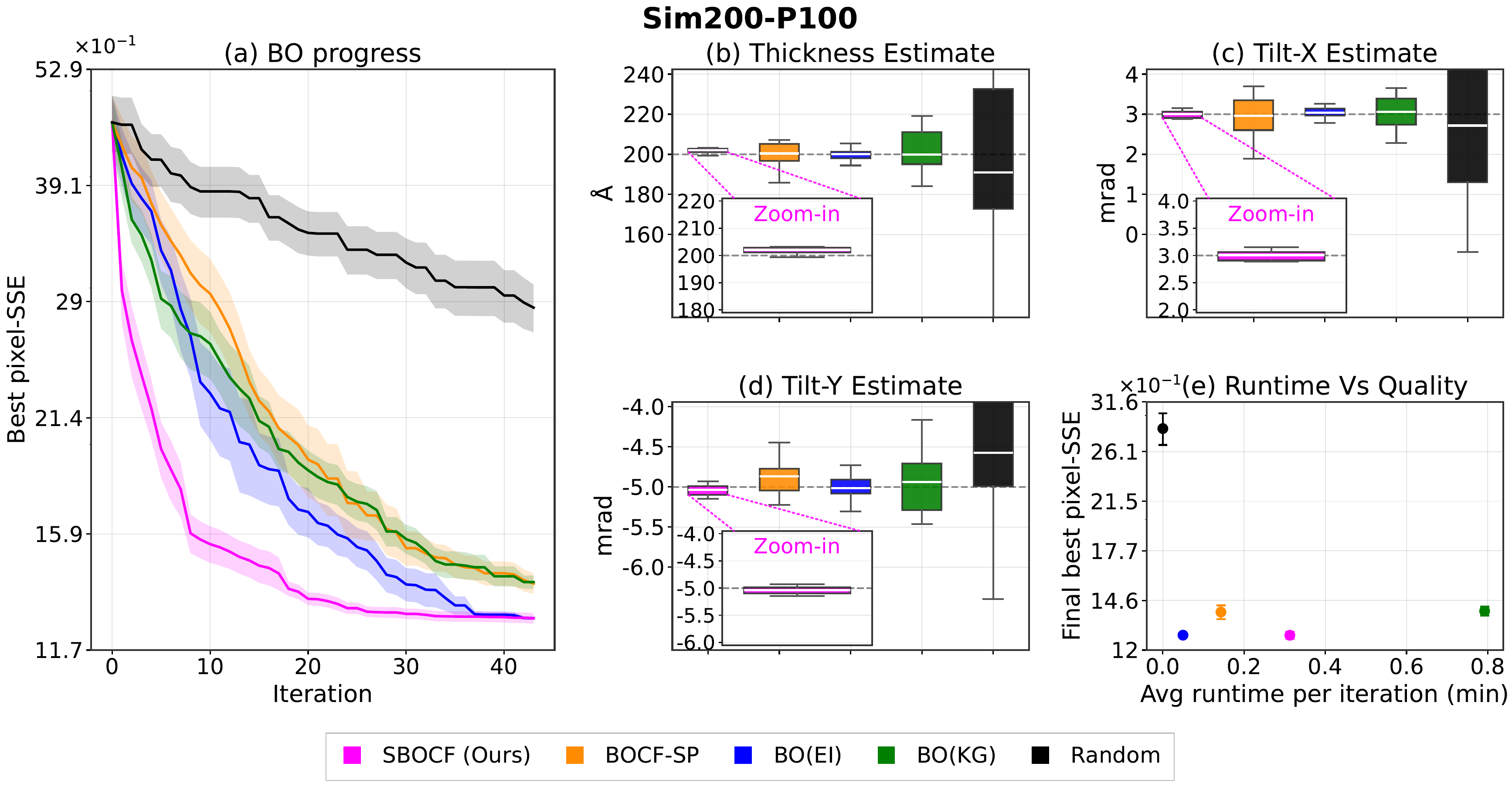}
\caption{
Results for \texttt{Sim200-P100} with ground-truth thickness $200$~\AA{}, tilts $(3,-5)$~mrad, and Poisson noise with a peak photon count of $100$. Panels show (a) best observed pixel-wise SSE, final estimates of (b) thickness, (c) tilt-X, and (d) tilt-Y, and (e) acquisition runtime versus final SSE. Solid lines in Panel~(a) show the mean across 20 trials, and the shaded regions indicate standard error. Dashed lines in Panels~(b)--(d) indicate the ground truth.
}
    \label{fig:sim200-p100}
\end{figure}

\begin{figure}[h!]
    \centering
\includegraphics[width=0.9\linewidth]{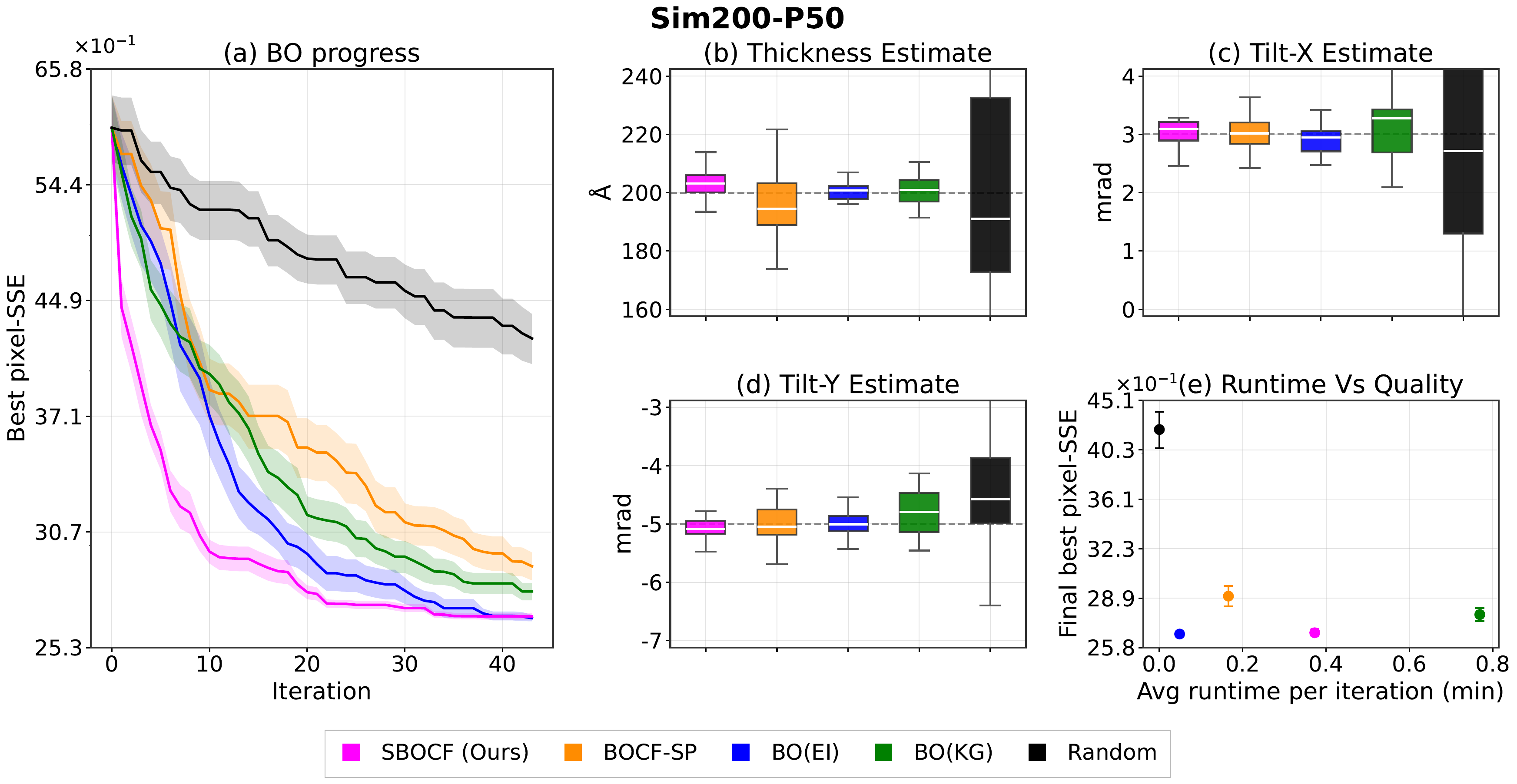}
\caption{
Results for \texttt{Sim200-P50} with ground-truth thickness $200$~\AA{}, tilts $(3,-5)$~mrad, and Poisson noise with a peak photon count of $50$. Panels show (a) best observed pixel-wise SSE, final estimates of (b) thickness, (c) tilt-X, and (d) tilt-Y, and (e) acquisition runtime versus final SSE. Solid lines in Panel~(a) show the mean across 20 trials, and the shaded regions indicate standard error. Dashed lines in Panels~(b)--(d) indicate the ground truth.
}
    \label{fig:sim200-p50}
\end{figure}

Figs.~\ref{fig:sim380-p150}--\ref{fig:sim380-p50},
\ref{fig:sim100-p150}--\ref{fig:sim100-p50}, and
\ref{fig:sim200-p150}--\ref{fig:sim200-p50} summarize the noisy \texttt{Sim380}, \texttt{Sim100}, and \texttt{Sim200} results, respectively, for peak photon counts of 150, 100, and 50. Each figure shows (a) the best observed pixel-wise SSE, (b)--(d) the final thickness, tilt-X, and tilt-Y estimates, and (e) acquisition runtime versus final SSE. Dashed lines in panels (b)--(d) indicate the ground-truth parameter values.

Overall, we observe that the benefits of leveraging composite structure become less pronounced as the observation noise increases. In particular, under the highest noise setting, the performance gap between SBOCF and standard BO methods narrows substantially, especially for the thinner-sample \texttt{Sim100} problems. This behavior is expected since heavy pixel-level noise partially obscures the underlying composite structure exploited by SBOCF. Nevertheless, SBOCF remains consistently competitive across all noise levels and continues to provide accurate parameter estimates in many settings.

The results for BOCF-SP are consistent with those reported in the main
text: it underperforms SBOCF and is comparable to, or slightly worse
than, standard BO. As discussed in Section~\ref{sec:results}, BOCF-SP
does not fully exploit the patch-level composite structure as SBOCF
does. Instead, it models information similar to that used by standard
BO through a more complicated surrogate, introducing opportunities
for modeling error to propagate without a commensurate gain in useful
information.

We additionally observe that the thinner-sample cases are substantially more sensitive to noise than the thicker-sample cases. In the \texttt{Sim380} experiments, SBOCF continues to provide clear improvements in both optimization performance and parameter estimation consistency even under noisy conditions. In contrast, for the thinner-sample  \texttt{Sim100} and \texttt{Sim200} problems, the optimization quality and parameter estimation consistency degrade more noticeably as the noise level increases. This observation is consistent with a prior PACBED study~\cite{xu_deep_2018}, which suggests that PACBED patterns contain more reliable structural information for samples thicker than 200~\textup{\AA}, leading to more stable parameter recovery.



\clearpage
\section{Ablation study: Alternative partition pattern}\label{app:newpattern}

In this section, we conduct an additional ablation study to investigate the effect of the image partitioning pattern used in SBOCF. In the main experiments, we use the \texttt{square} pattern shown in Panel~(b) of Fig.~\ref{fig:patch_pixel_sse}, where the PACBED image is divided into a regular grid of square patches. Here, we compare this choice with an alternative \texttt{domain} pattern motivated by the intensity structure of PACBED images.

\begin{figure}[h!]
    \centering
    \includegraphics[width=0.5\linewidth]{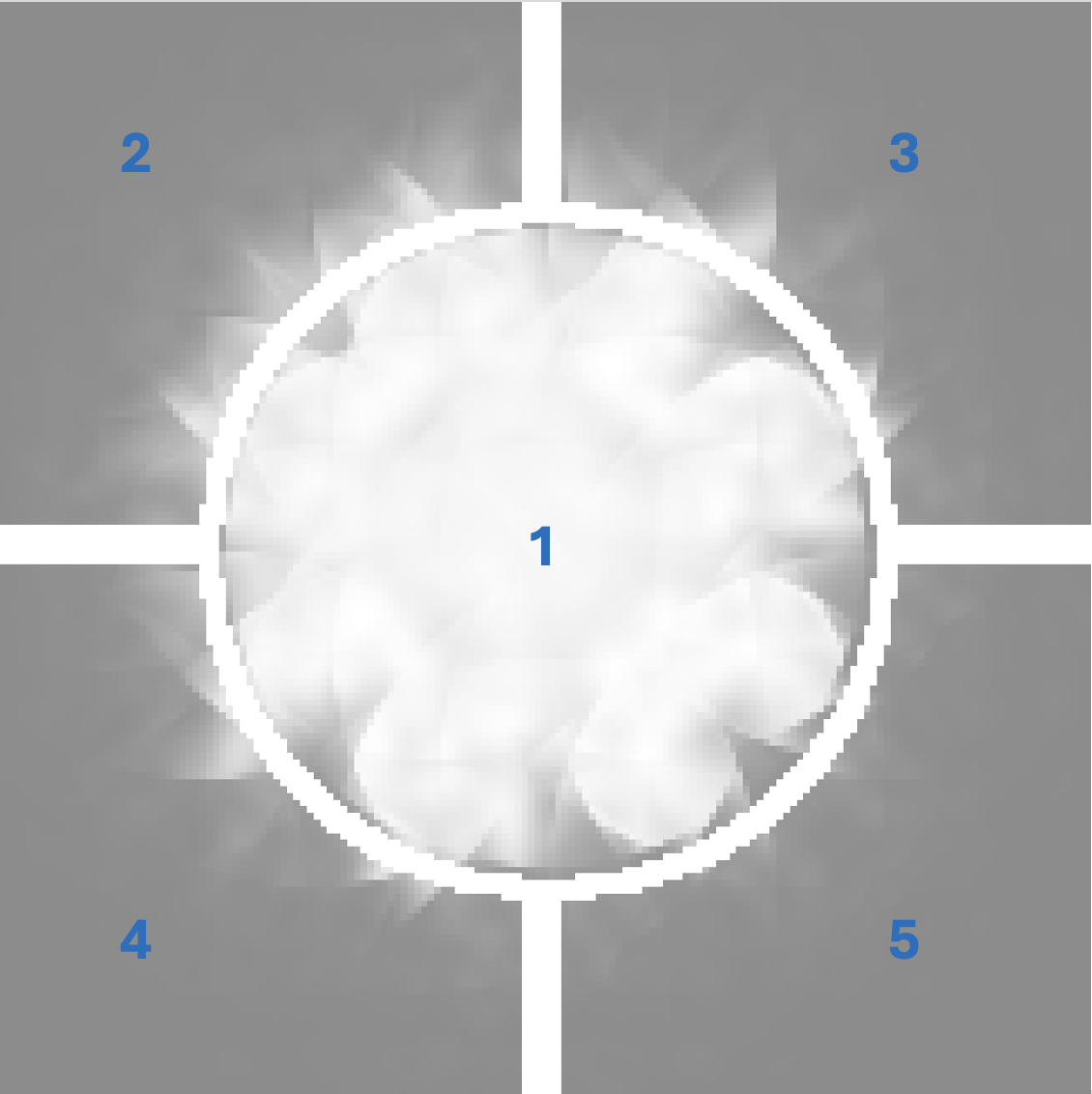}
    \caption{Illustration of the \texttt{domain} PACBED segmentation pattern used in the ablation study. The image is partitioned into five regions: one central circular region containing high-intensity pixels and four surrounding quadrant regions containing lower-intensity pixels.}
    \label{fig:domain_pattern}
\end{figure}

The \texttt{domain} pattern partitions the PACBED image into five regions: one circular region containing the high-intensity pixels near the center of the image, and four outer quadrant regions containing lower-intensity pixels. This pattern is illustrated in Fig.~\ref{fig:domain_pattern}. We refer to this partition as the \texttt{domain} pattern because it separates the image into broad intensity-based regions rather than regular spatial patches.

\begin{figure}
    \centering
    \includegraphics[width=0.6\linewidth]{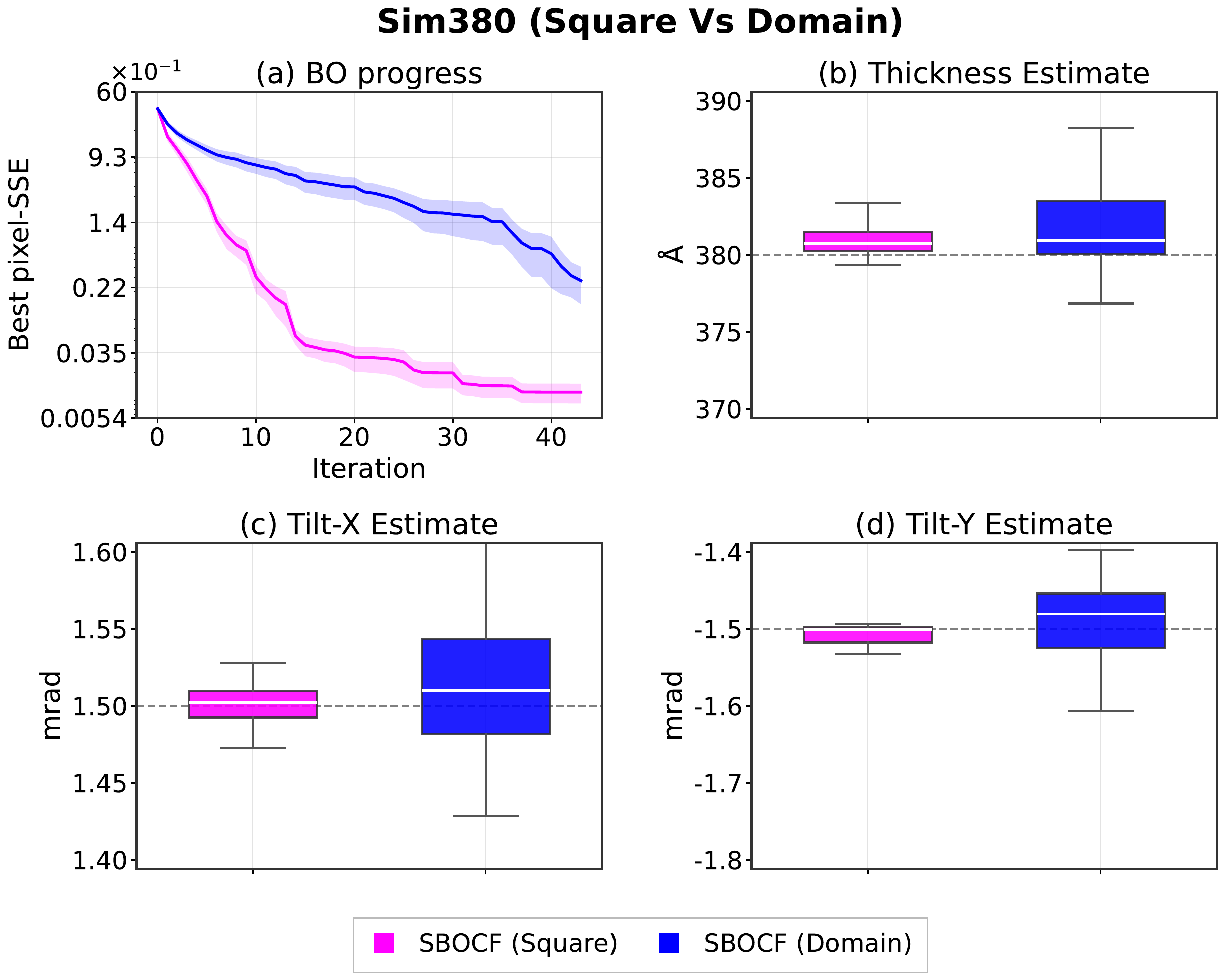}
    \caption{Comparison of SBOCF using the \texttt{square} and \texttt{domain} partition patterns on the noiseless \texttt{Sim380} problem with ground-truth thickness $380$~\AA{} and tilts $(1.5,-1.5)$~mrad. Panels show (a) best observed pixel-wise SSE and final estimates of (b) thickness, (c) tilt-X, and (d) tilt-Y. Solid lines in Panel~(a) show the mean across 20 trials, and the shaded regions indicate standard error. Dashed lines in Panels~(b)--(d) indicate the ground truth.}
    \label{fig:sim380-sqvsdm}
\end{figure}

\begin{figure}
    \centering
    \includegraphics[width=0.6\linewidth]{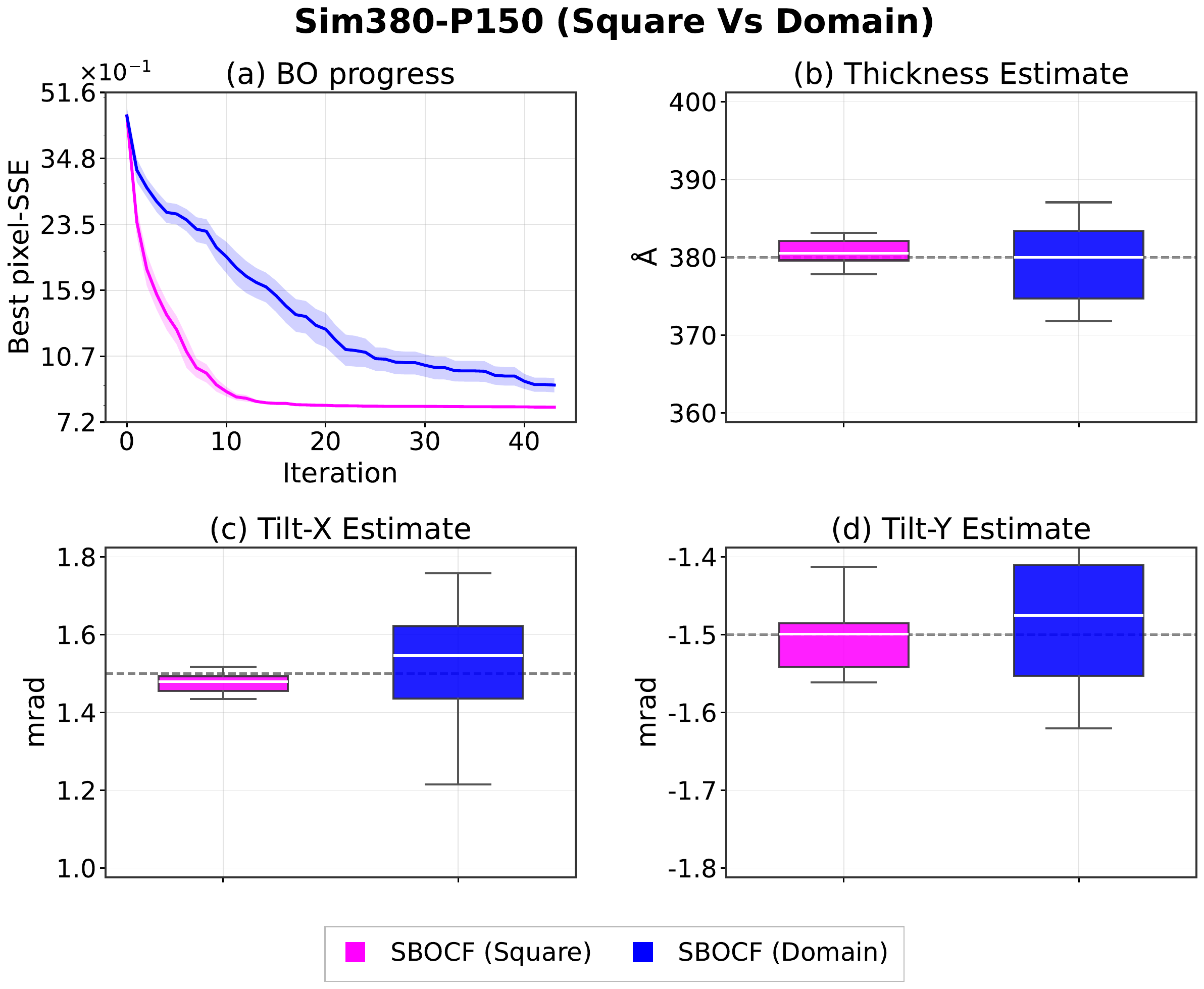}
    \caption{Comparison of SBOCF using the \texttt{square} and \texttt{domain} partition patterns on the \texttt{Sim380-P150} problem with ground-truth thickness $380$~\AA{}, tilts $(1.5,-1.5)$~mrad and Poisson noise with a peak photon count of 150. Panels show (a) best observed pixel-wise SSE and final estimates of (b) thickness, (c) tilt-X, and (d) tilt-Y. Solid lines in Panel~(a) show the mean across 20 trials, and the shaded regions indicate standard error. Dashed lines in Panels~(b)--(d) indicate the ground truth.}
    \label{fig:sim380-p150-sqvsdm}
\end{figure}

\begin{figure}
    \centering
    \includegraphics[width=0.6\linewidth]{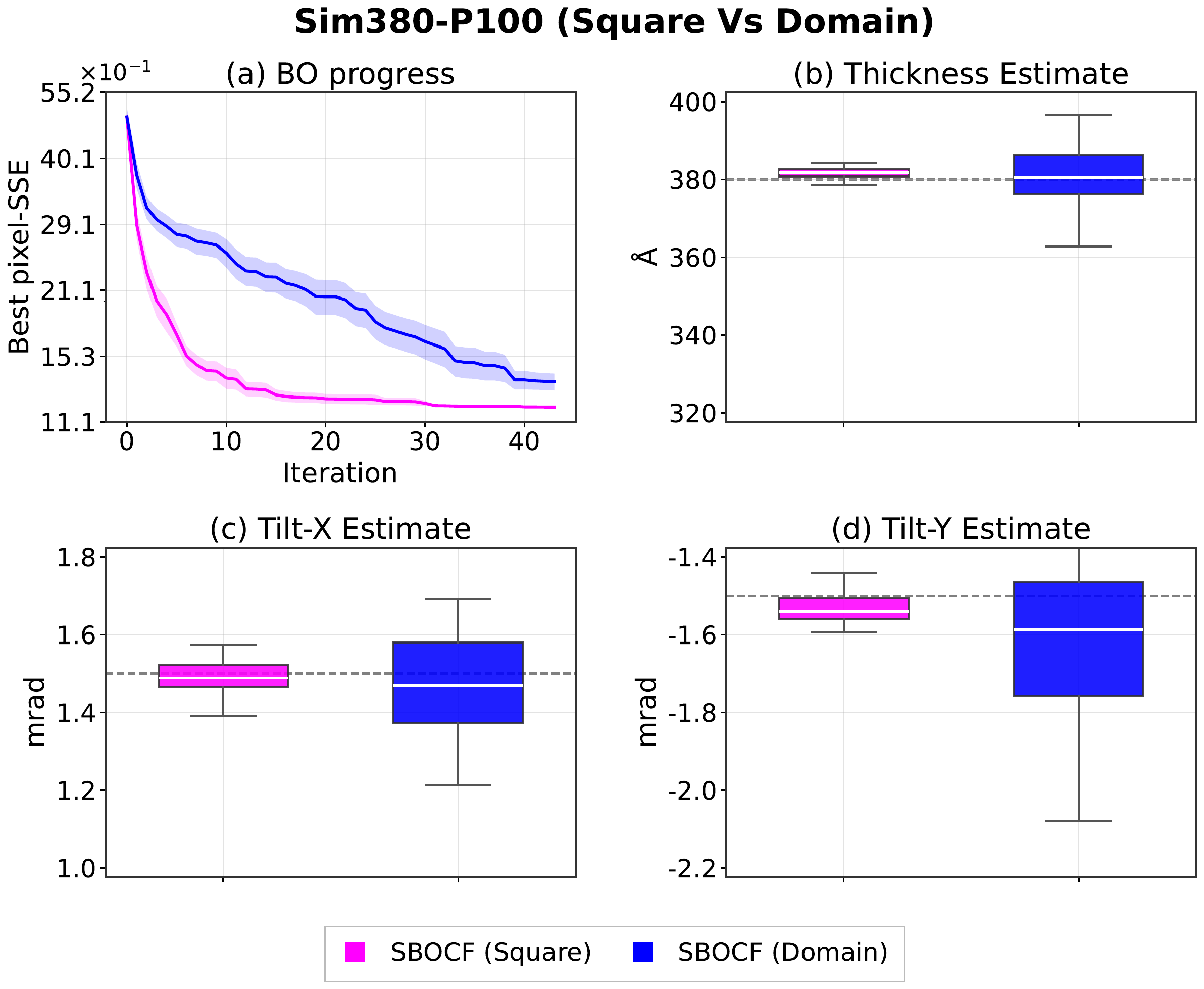}
    \caption{Comparison of SBOCF using the \texttt{square} and \texttt{domain} partition patterns on the \texttt{Sim380-P100} problem with ground-truth thickness $380$~\AA{}, tilts $(1.5,-1.5)$~mrad and Poisson noise with a peak photon count of 100. Panels show (a) best observed pixel-wise SSE and final estimates of (b) thickness, (c) tilt-X, and (d) tilt-Y. Solid lines in Panel~(a) show the mean across 20 trials, and the shaded regions indicate standard error. Dashed lines in Panels~(b)--(d) indicate the ground truth.}
    \label{fig:sim380-p100-sqvsdm}
\end{figure}

\begin{figure}
    \centering
    \includegraphics[width=0.6\linewidth]{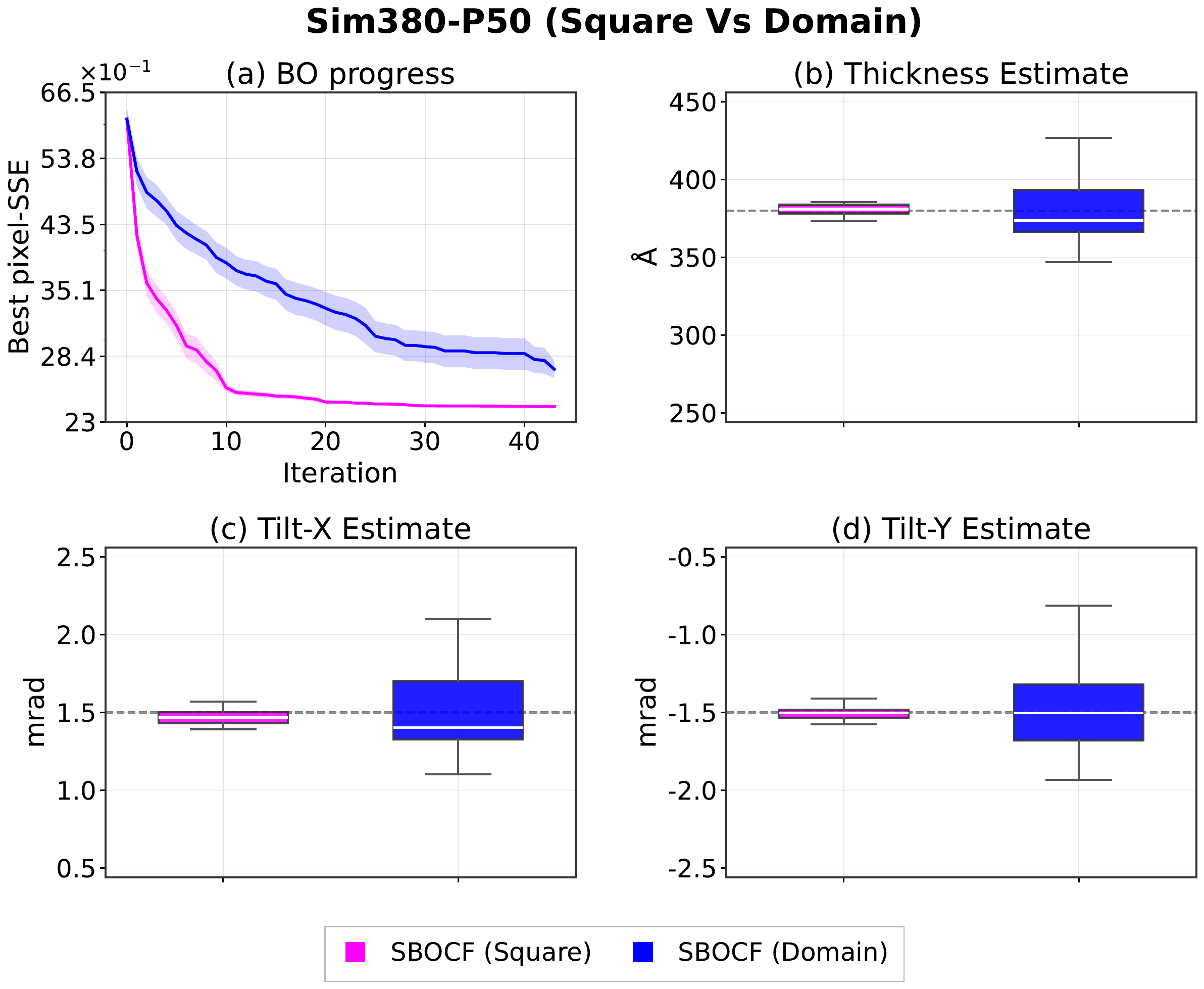}
    \caption{Comparison of SBOCF using the \texttt{square} and \texttt{domain} partition patterns on the \texttt{Sim380-P50} problem with ground-truth thickness $380$~\AA{}, tilts $(1.5,-1.5)$~mrad and Poisson noise with a peak photon count of 50. Panels show (a) best observed pixel-wise SSE and final estimates of (b) thickness, (c) tilt-X, and (d) tilt-Y. Solid lines in Panel~(a) show the mean across 20 trials, and the shaded regions indicate standard error. Dashed lines in Panels~(b)--(d) indicate the ground truth.}
    \label{fig:sim380-p50-sqvsdm}
\end{figure}

\begin{figure}
    \centering
    \includegraphics[width=0.6\linewidth]{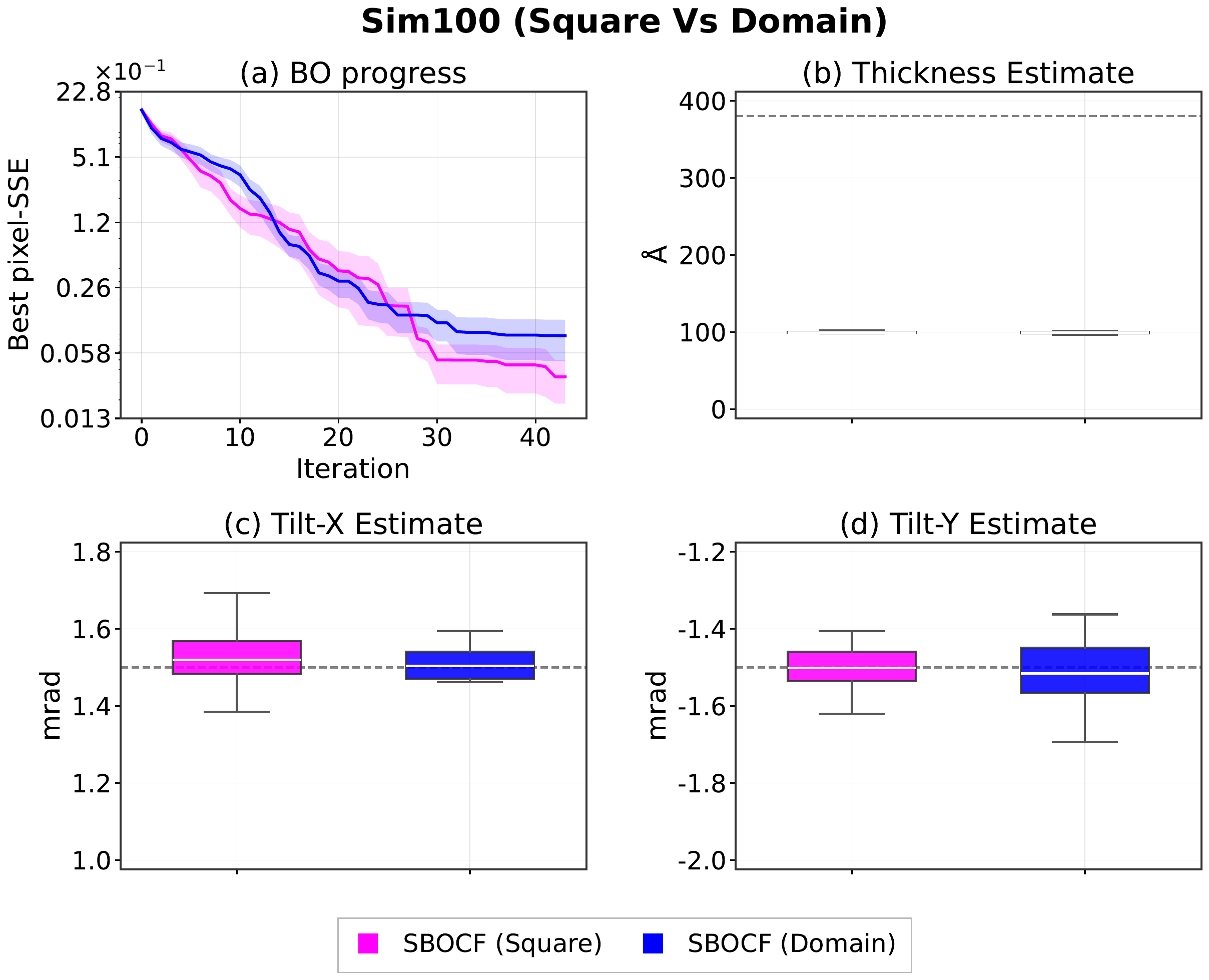}
    \caption{Comparison of SBOCF using the \texttt{square} and \texttt{domain} partition patterns on the noiseless \texttt{Sim100} problem with ground-truth thickness $100$~\AA{} and tilts $(1.5,-1.5)$~mrad. Panels show (a) best observed pixel-wise SSE and final estimates of (b) thickness, (c) tilt-X, and (d) tilt-Y. Solid lines in Panel~(a) show the mean across 20 trials, and the shaded regions indicate standard error. Dashed lines in Panels~(b)--(d) indicate the ground truth.}
    \label{fig:sim100-sqvsdm}
\end{figure}

\begin{figure}
    \centering
    \includegraphics[width=0.6\linewidth]{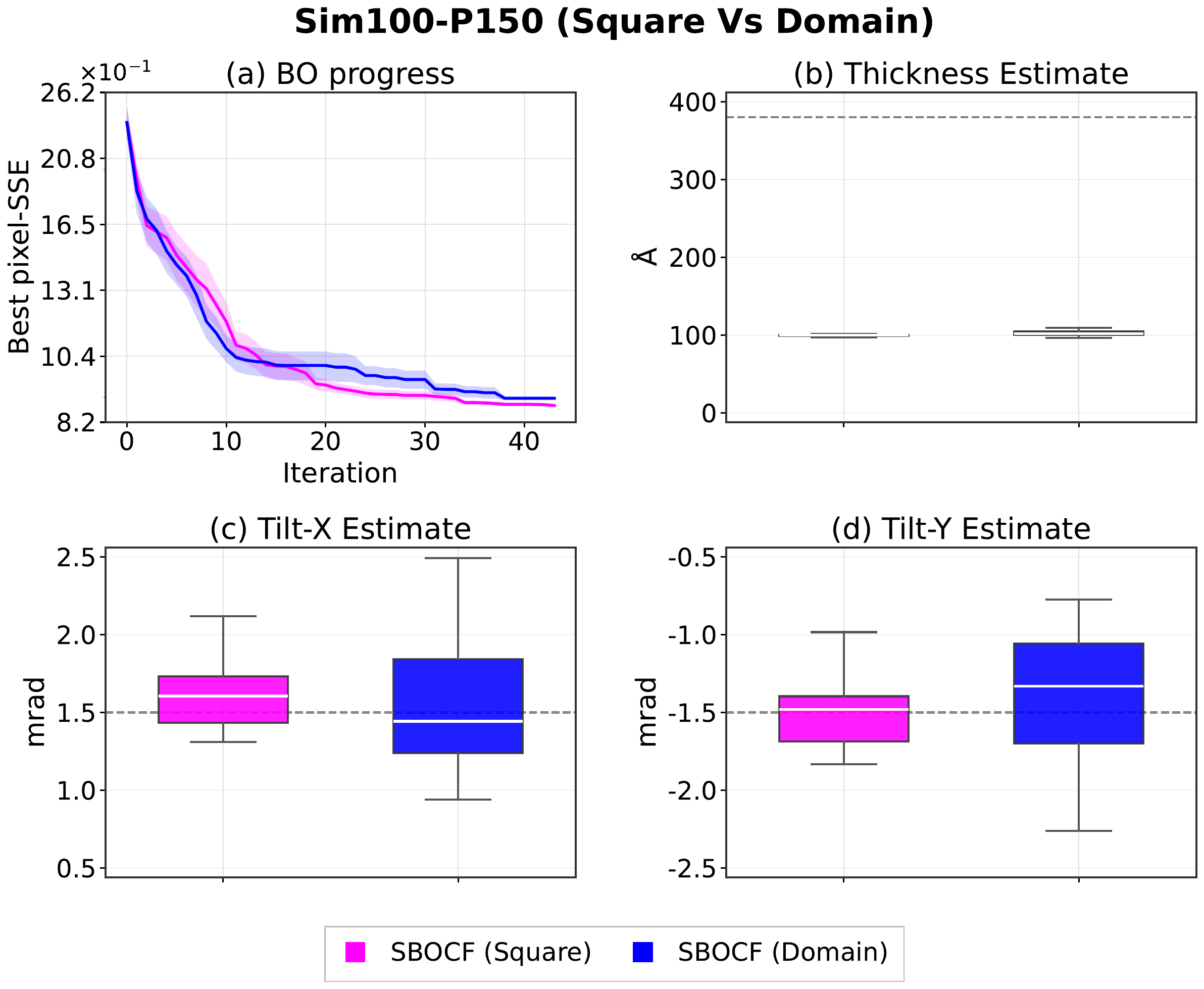}
    \caption{Comparison of SBOCF using the \texttt{square} and \texttt{domain} partition patterns on the \texttt{Sim100-P150} problem with ground-truth thickness $100$~\AA{}, tilts $(1.5,-1.5)$~mrad and Poisson noise with a peak photon count of 150. Panels show (a) best observed pixel-wise SSE and final estimates of (b) thickness, (c) tilt-X, and (d) tilt-Y. Solid lines in Panel~(a) show the mean across 20 trials, and the shaded regions indicate standard error. Dashed lines in Panels~(b)--(d) indicate the ground truth.}
    \label{fig:sim100-p150-sqvsdm}
\end{figure}

\begin{figure}
    \centering
    \includegraphics[width=0.6\linewidth]{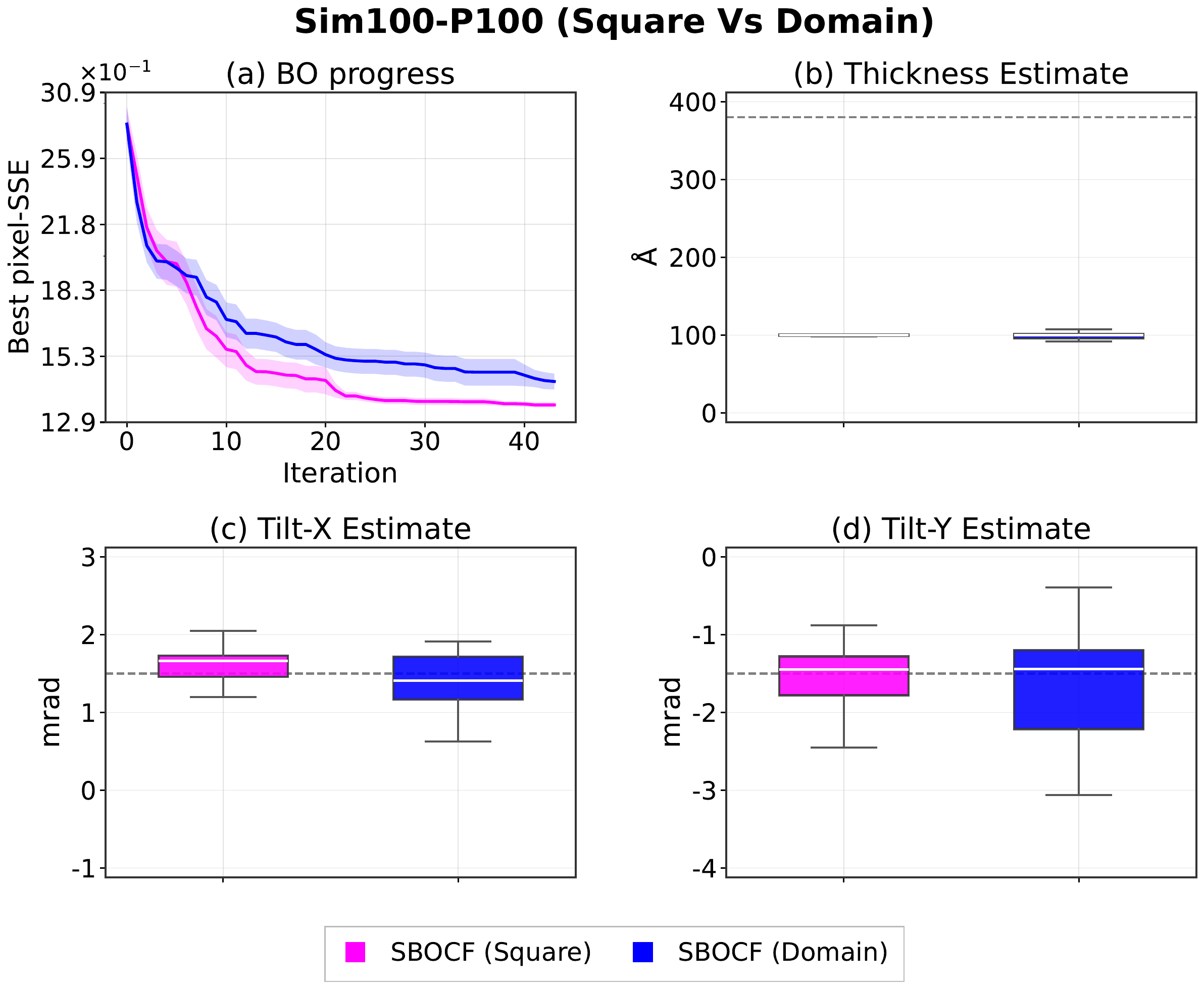}
    \caption{Comparison of SBOCF using the \texttt{square} and \texttt{domain} partition patterns on the \texttt{Sim100-P100} problem with ground-truth thickness $100$~\AA{}, tilts $(1.5,-1.5)$~mrad and Poisson noise with a peak photon count of 100. Panels show (a) best observed pixel-wise SSE and final estimates of (b) thickness, (c) tilt-X, and (d) tilt-Y. Solid lines in Panel~(a) show the mean across 20 trials, and the shaded regions indicate standard error. Dashed lines in Panels~(b)--(d) indicate the ground truth.}
    \label{fig:sim100-p100-sqvsdm}
\end{figure}

\begin{figure}
    \centering
    \includegraphics[width=0.6\linewidth]{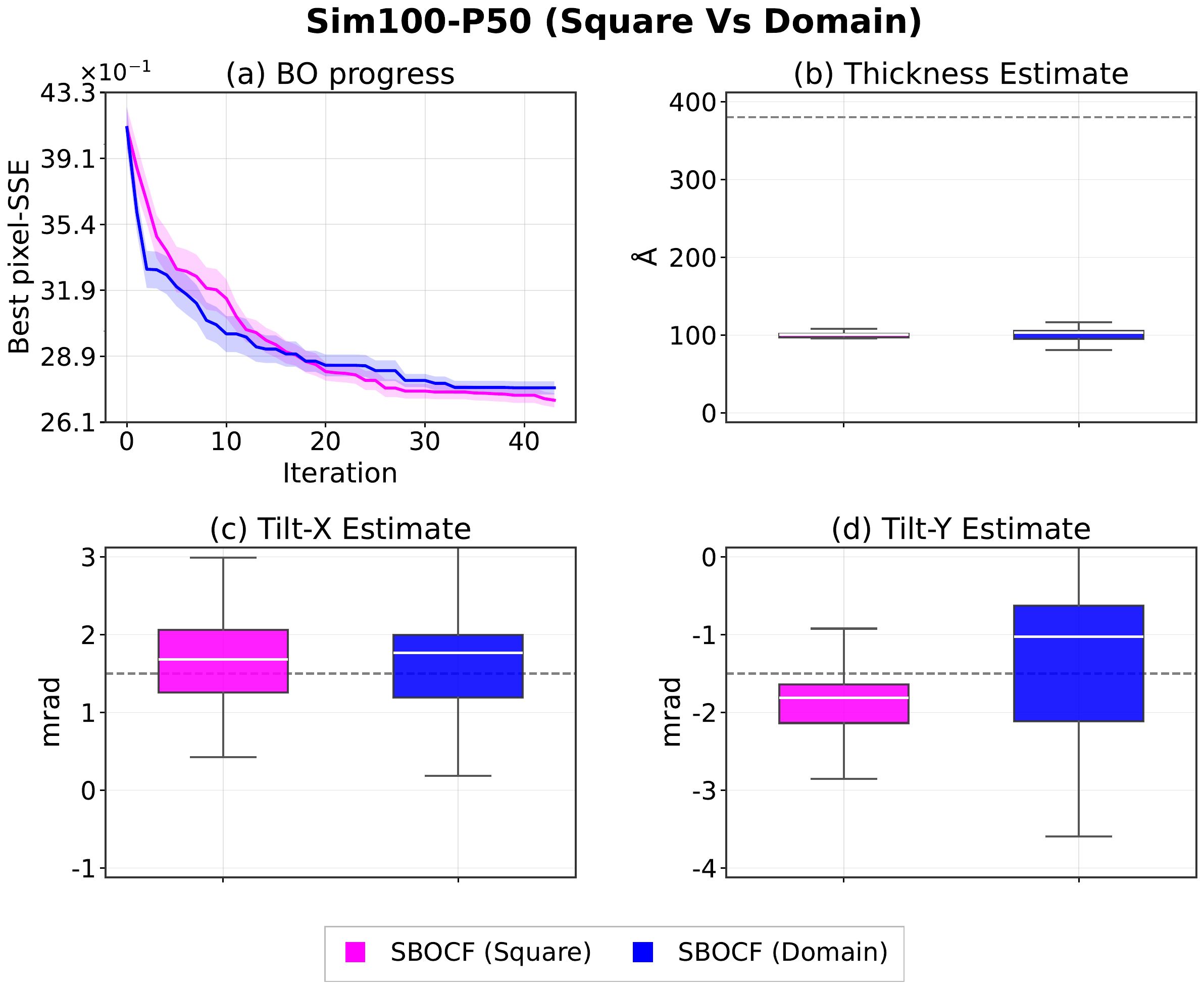}
    \caption{Comparison of SBOCF using the \texttt{square} and \texttt{domain} partition patterns on the \texttt{Sim100-P50} problem with ground-truth thickness $100$~\AA{}, tilts $(1.5,-1.5)$~mrad and Poisson noise with a peak photon count of 50. Panels show (a) best observed pixel-wise SSE and final estimates of (b) thickness, (c) tilt-X, and (d) tilt-Y. Solid lines in Panel~(a) show the mean across 20 trials, and the shaded regions indicate standard error. Dashed lines in Panels~(b)--(d) indicate the ground truth.}
    \label{fig:sim100-p50-sqvsdm}
\end{figure}

\begin{figure}
    \centering
    \includegraphics[width=0.6\linewidth]{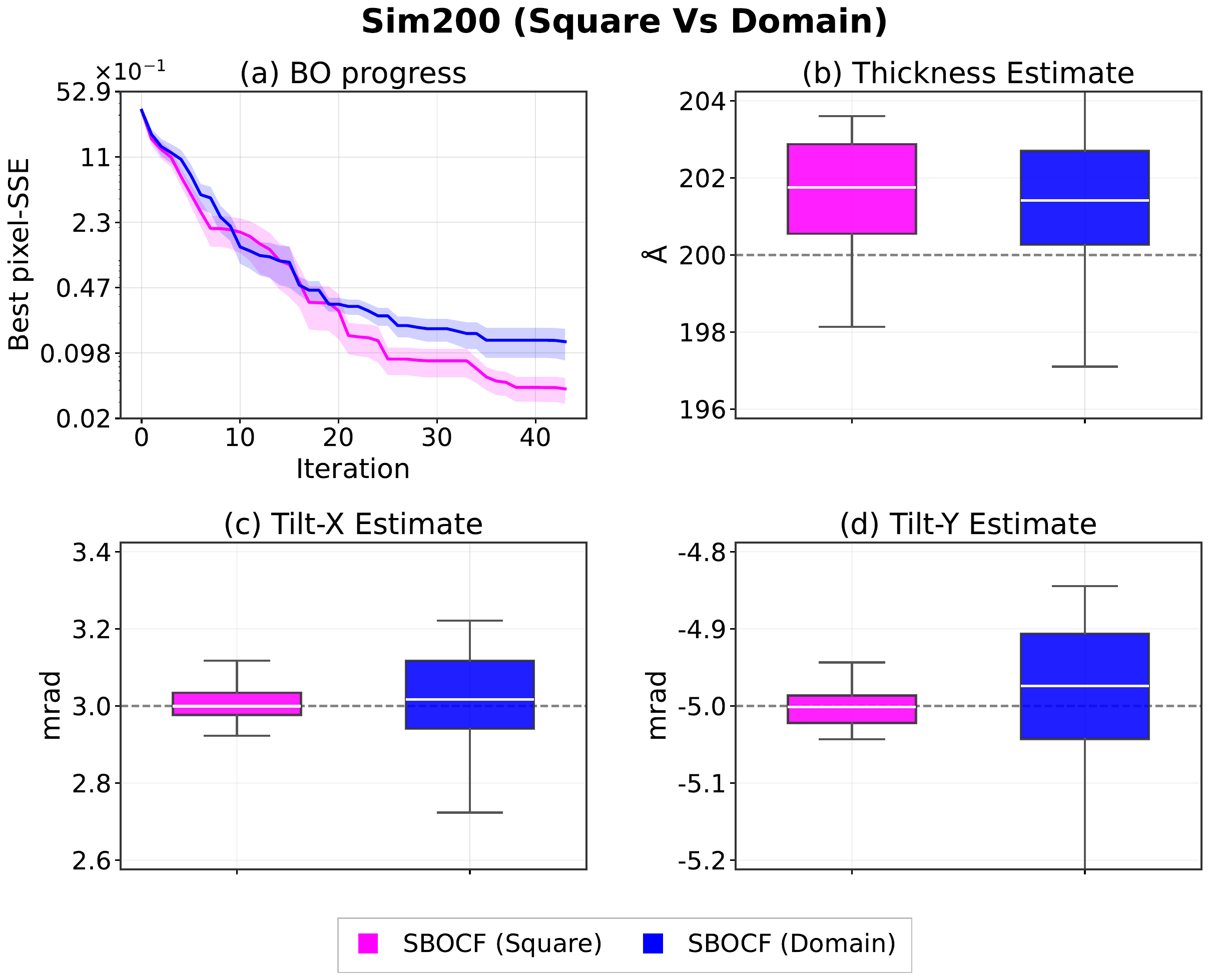}
    \caption{Comparison of SBOCF using the \texttt{square} and \texttt{domain} partition patterns on the noiseless \texttt{Sim200} problem with ground-truth thickness $200$~\AA{} and tilts $(3,-5)$~mrad. Panels show (a) best observed pixel-wise SSE and final estimates of (b) thickness, (c) tilt-X, and (d) tilt-Y. Solid lines in Panel~(a) show the mean across 20 trials, and the shaded regions indicate standard error. Dashed lines in Panels~(b)--(d) indicate the ground truth.}
    \label{fig:sim200-sqvsdm}
\end{figure}

\begin{figure}
    \centering
    \includegraphics[width=0.6\linewidth]{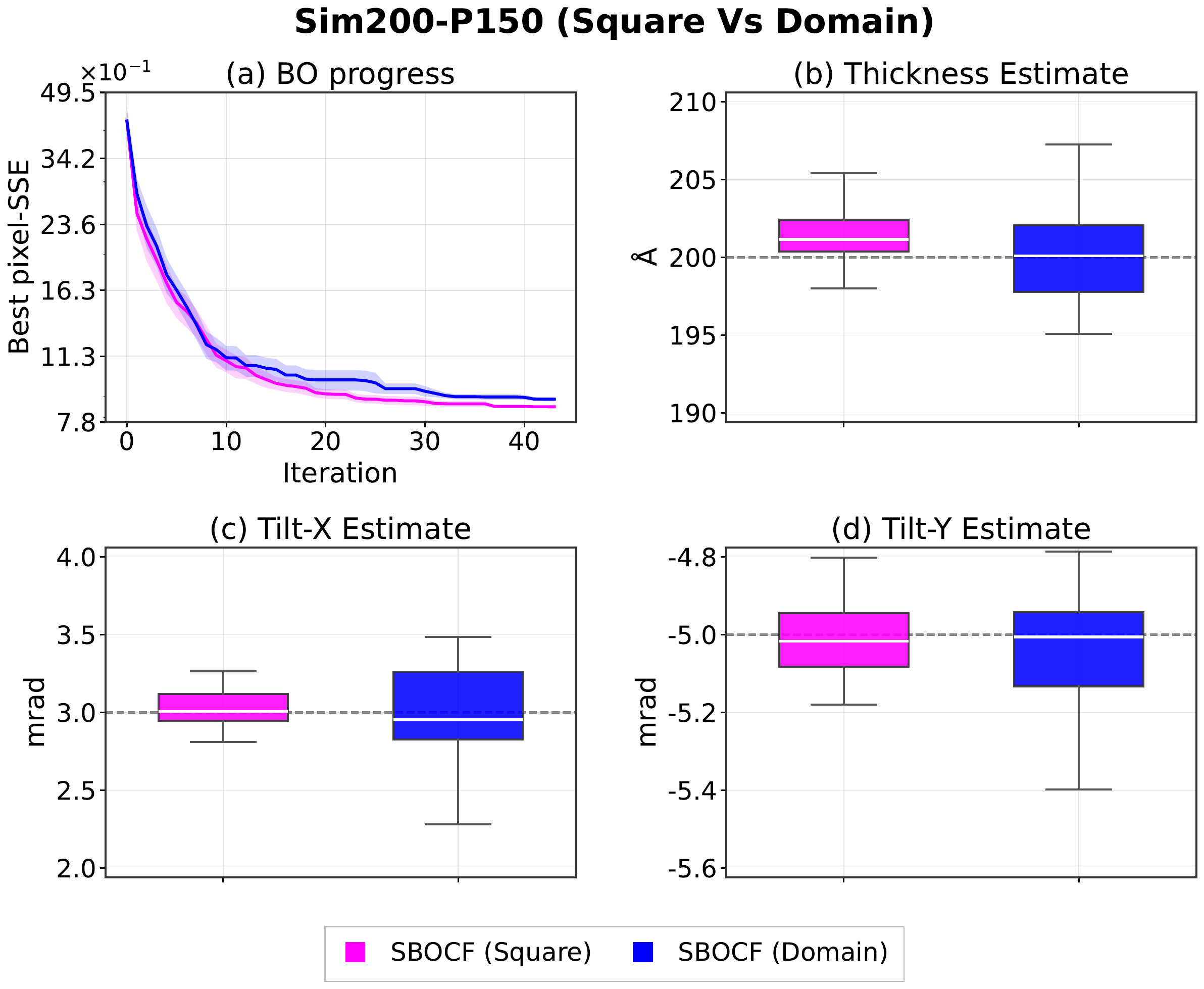}
    \caption{Comparison of SBOCF using the \texttt{square} and \texttt{domain} partition patterns on the \texttt{Sim200-P150} problem with ground-truth thickness $200$~\AA{}, tilts $(3,-5)$~mrad and Poisson noise with a peak photon count of 150. Panels show (a) best observed pixel-wise SSE and final estimates of (b) thickness, (c) tilt-X, and (d) tilt-Y. Solid lines in Panel~(a) show the mean across 20 trials, and the shaded regions indicate standard error. Dashed lines in Panels~(b)--(d) indicate the ground truth.}
    \label{fig:sim200-p150-sqvsdm}
\end{figure}

\begin{figure}
    \centering
    \includegraphics[width=0.6\linewidth]{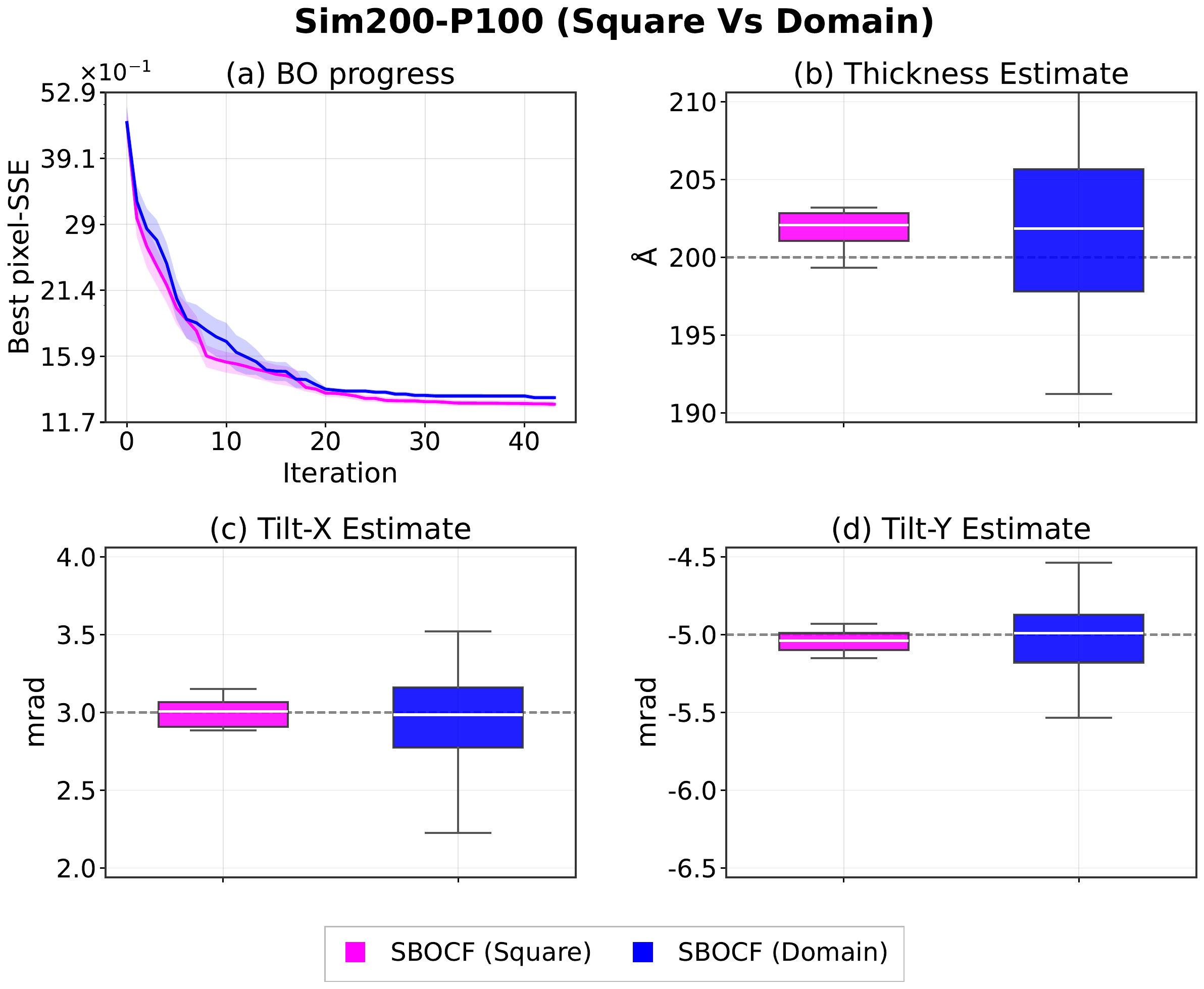}
    \caption{Comparison of SBOCF using the \texttt{square} and \texttt{domain} partition patterns on the \texttt{Sim200-P100} problem with ground-truth thickness $200$~\AA{}, tilts $(3,-5)$~mrad and Poisson noise with a peak photon count of 100. Panels show (a) best observed pixel-wise SSE and final estimates of (b) thickness, (c) tilt-X, and (d) tilt-Y. Solid lines in Panel~(a) show the mean across 20 trials, and the shaded regions indicate standard error. Dashed lines in Panels~(b)--(d) indicate the ground truth.}
    \label{fig:sim200-p100-sqvsdm}
\end{figure}

\begin{figure}
    \centering
    \includegraphics[width=0.6\linewidth]{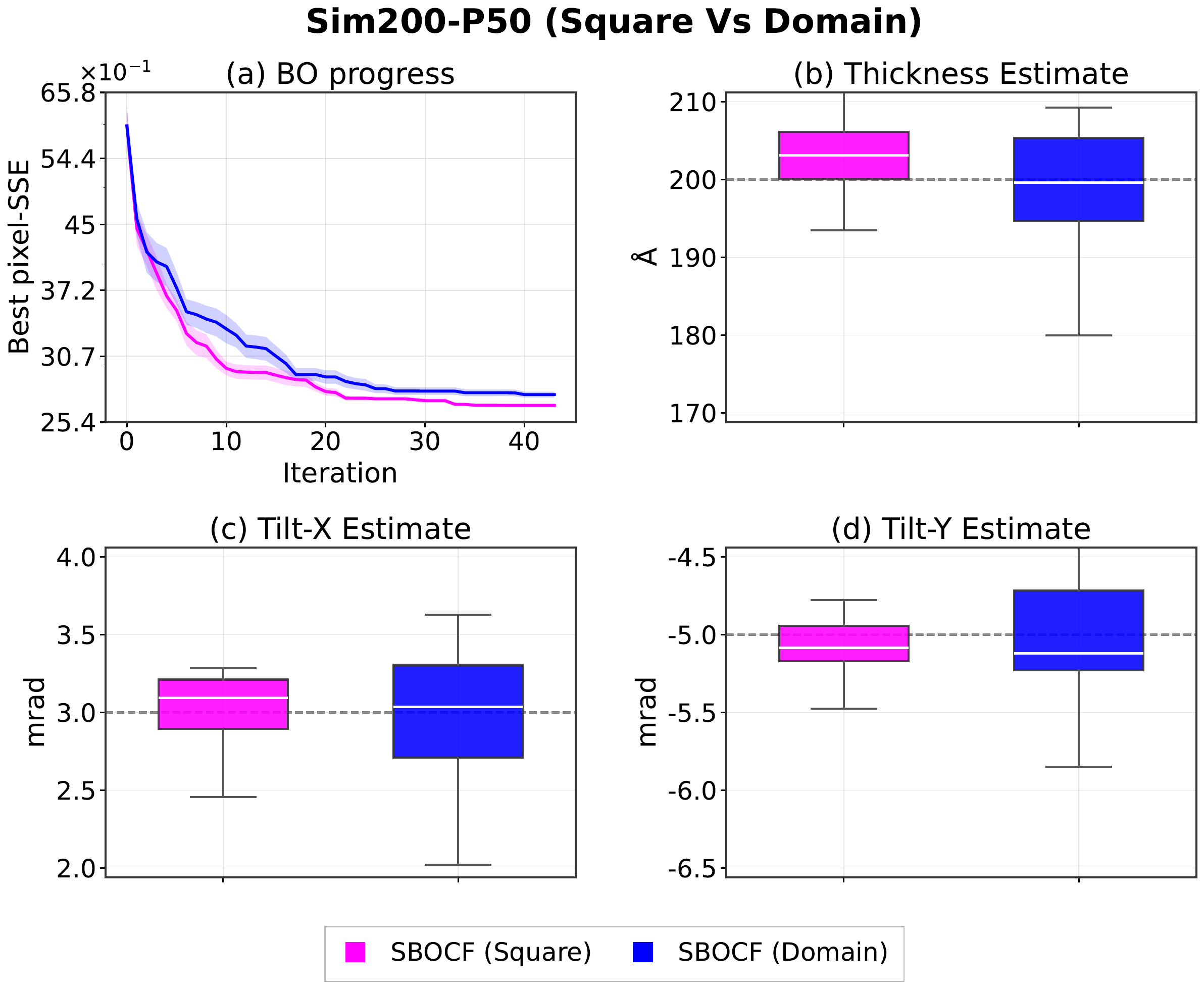}
    \caption{Comparison of SBOCF using the \texttt{square} and \texttt{domain} partition patterns on the \texttt{Sim200-P50} problem with ground-truth thickness $200$~\AA{}, tilts $(3,-5)$~mrad and Poisson noise with a peak photon count of 50. Panels show (a) best observed pixel-wise SSE and final estimates of (b) thickness, (c) tilt-X, and (d) tilt-Y. Solid lines in Panel~(a) show the mean across 20 trials, and the shaded regions indicate standard error. Dashed lines in Panels~(b)--(d) indicate the ground truth.}
    \label{fig:sim200-p50-sqvsdm}
\end{figure}

We perform this comparison on all three synthetic problems, \texttt{Sim380}, \texttt{Sim100}, and \texttt{Sim200}, under both noiseless and noisy settings. For the noisy cases, we consider peak photon counts \texttt{P150}, \texttt{P100}, and \texttt{P50}. Figs.~\ref{fig:sim380-sqvsdm}--\ref{fig:sim380-p50-sqvsdm}, \ref{fig:sim100-sqvsdm}--\ref{fig:sim100-p50-sqvsdm}, and \ref{fig:sim200-sqvsdm}--\ref{fig:sim200-p50-sqvsdm} show the results for \texttt{Sim380}, \texttt{Sim100}, and \texttt{Sim200}, respectively. In each figure, Panel~(a) shows the best observed pixel-wise SSE, while Panels~(b)--(d) show the final thickness, tilt-X, and tilt-Y estimates.

Overall, the \texttt{square} pattern performs better than, or at least comparably to, the \texttt{domain} pattern across the tested settings. The advantage of the \texttt{square} pattern is especially clear for \texttt{Sim380}, where it achieves faster reduction in pixel-SSE and produces more stable final parameter estimates under both noiseless and noisy observations. For thinner problems, e.g. \texttt{Sim100} and \texttt{Sim200}, the difference is smaller in some settings, but the \texttt{square} pattern still generally gives comparable or lower final pixel-SSE and more stable parameter estimates.

One possible explanation is that the \texttt{domain} pattern groups pixels primarily according to broad high- and low-intensity regions. Although this choice is physically motivated, it may be too coarse for this parameter estimation problem and may discard spatial information that is useful for distinguishing thickness and tilt effects in PACBED images. In contrast, the \texttt{square} pattern preserves more local spatial information by dividing the image into smaller regular patches. These results suggest that the choice of partition pattern can affect the performance of SBOCF, and designing problem-adaptive partition patterns is worth further investigation.

%%%%%%%%%%%%%%%%%%%%%%%%%%%%%%%%%%%%%%%%%%%%%%%%%%%%%%%%%%%%

% \newpage
% \input{checklist.tex}

\end{document}